\documentclass[final,5p,times]{elsarticle}

\usepackage{caption} 
\usepackage{cuted}   

\usepackage[utf8]{inputenc}
\usepackage[T1]{fontenc}
\usepackage{amsmath}
\usepackage{amssymb}
\usepackage{amsfonts}
\usepackage{mathtools}
\usepackage{amsthm}
\usepackage{bm}
\usepackage{booktabs}
\usepackage{graphicx}
\usepackage{microtype}
\usepackage{nicefrac}
\usepackage{url}
\usepackage{xcolor}
\usepackage{algorithm, algorithmic}
\usepackage{dashrule}
\usepackage{subfigure}
\usepackage{siunitx}
\usepackage{multirow}
\usepackage{pifont}
\newcommand{\cmark}{\ding{51}}
\newcommand{\xmark}{\ding{55}}

\theoremstyle{plain}
\newtheorem{theorem}{Theorem}

\newtheorem{lemma}{Lemma}

\theoremstyle{definition}

\newtheorem{assumption}{Assumption}
\theoremstyle{remark}
\newtheorem{remark}{Remark}

\begin{document}

\begin{frontmatter}



\title{Communication-Efficient Personalized Federated Learning via Layer-Wise Multi-Threshold Random Sketching}


\author[1]{Xu Zhang}
\ead{zhang.xu@xidian.edu.cn}

\author[1]{Xingyu Hou}
\ead{24209100207@stu.xidian.edu.cn}

\author[2]{Jiacheng Cheng}
\ead{chengjc@mail.nwpu.edu.cn}

\author[3,4]{Kaiyuan Feng}
\ead{fkylwl@gmail.com}

\author[3,5]{Maoguo Gong\corref{cor1}}
\ead{gong@ieee.org}
\cortext[cor1]{Corresponding author}


\affiliation[1]{organization={School of Artificial Intelligence, Xidian University},
            city={Xi'an},
            postcode={710126},
            country={China}}
\affiliation[2]{organization={School of Automation, Northwestern Polytechnical University},
            city={Xi'an},
            country={China}}
\affiliation[3]{organization={Key Laboratory of Collaborative Intelligence Systems, Ministry of Education, Xidian University},
            city={Xi'an},
            postcode={710071},
            country={China}}
\affiliation[4]{organization={Faculty of Infor-X, Xidian University},
            city={Xi'an},
            postcode={710071},
            country={China}}
\affiliation[5]{organization={College of Artificial Intelligence, Inner Mongolia Normal University},
            city={Hohhot},
            postcode={010022},
            country={China}}

\begin{abstract}
Personalized federated learning (PFL) is a promising paradigm for collaborative learning over distributed devices, where edge nodes collaboratively train personalized models without sharing raw data. Although PFL addresses data heterogeneity by learning client-specific models, it still suffers from substantial uplink and downlink communication costs when exchanging high-dimensional parameters in bandwidth-constrained systems. Recent one-bit methods achieve extreme compression, but they usually rely on a single thresholding rule applied to the whole model. This design has two limitations. First, it overlooks layer-wise differences in parameter distributions and quantization sensitivities. Second, a single threshold provides only coarse binary information and cannot capture fine-grained variations in parameter distributions. To address these issues, we propose a communication-efficient PFL framework via layer-wise multi-threshold random sketching. In the proposed method, each layer is assigned its own set of quantization thresholds, so that the compressed representation can adapt to layer-specific statistics while using multiple intervals to provide a finer low-bit description of sketched parameters. The proposed method supports bidirectional communication using compact low-bit sketches and improves the communication-accuracy tradeoff compared with existing one-bit compression approaches.
\end{abstract}

\begin{keyword}
Personalized federated learning \sep communication-efficient learning \sep  random sketching \sep multi-threshold quantization


\end{keyword}

\end{frontmatter}

\section{Introduction}

Federated learning enables collaborative model training across distributed devices while keeping raw data local, and has been widely considered in applications such as edge intelligence, internet of things, and vehicular networks~\cite{li2020federated2,kairouz2021advances,jia2025comprehensive,ZIAEINEJAD2026104446}. Personalized federated learning (PFL) further addresses data heterogeneity by learning client specific models rather than a single global model, which is useful when users, devices, locations, and sensing environments induce diverse local data distributions~\cite{dinh2020pfedme,zhang2022personalized,SABAH2024122874,ZHANG2025103221}. However, repeatedly exchanging high dimensional model parameters remains a fundamental communication bottleneck, especially in systems with limited bandwidth or a large number of participating clients~\cite{hamidi2025rate,cheng2026personalized}.
To mitigate this issue, recent studies have explored aggressive one-bit communication compression techniques based on sign-based aggregation, stochastic one-bit quantization, or one-bit random sketching, achieving substantial uplink and downlink communication savings \cite{cheng2026personalized,zhu2020obda,fan2022obcsaa,jin2022zsignfedavg,vargaftik2022eden,lan2025fedobma}. 
Despite their effectiveness in reducing communication overhead, these methods usually adopt layer-agnostic binary compression rules for model parameters or updates, which limits their expressiveness when applied to modern deep neural networks.

\begin{figure*}
    \centering
    \subfigure[Round 0]{\includegraphics[width=0.24\linewidth]{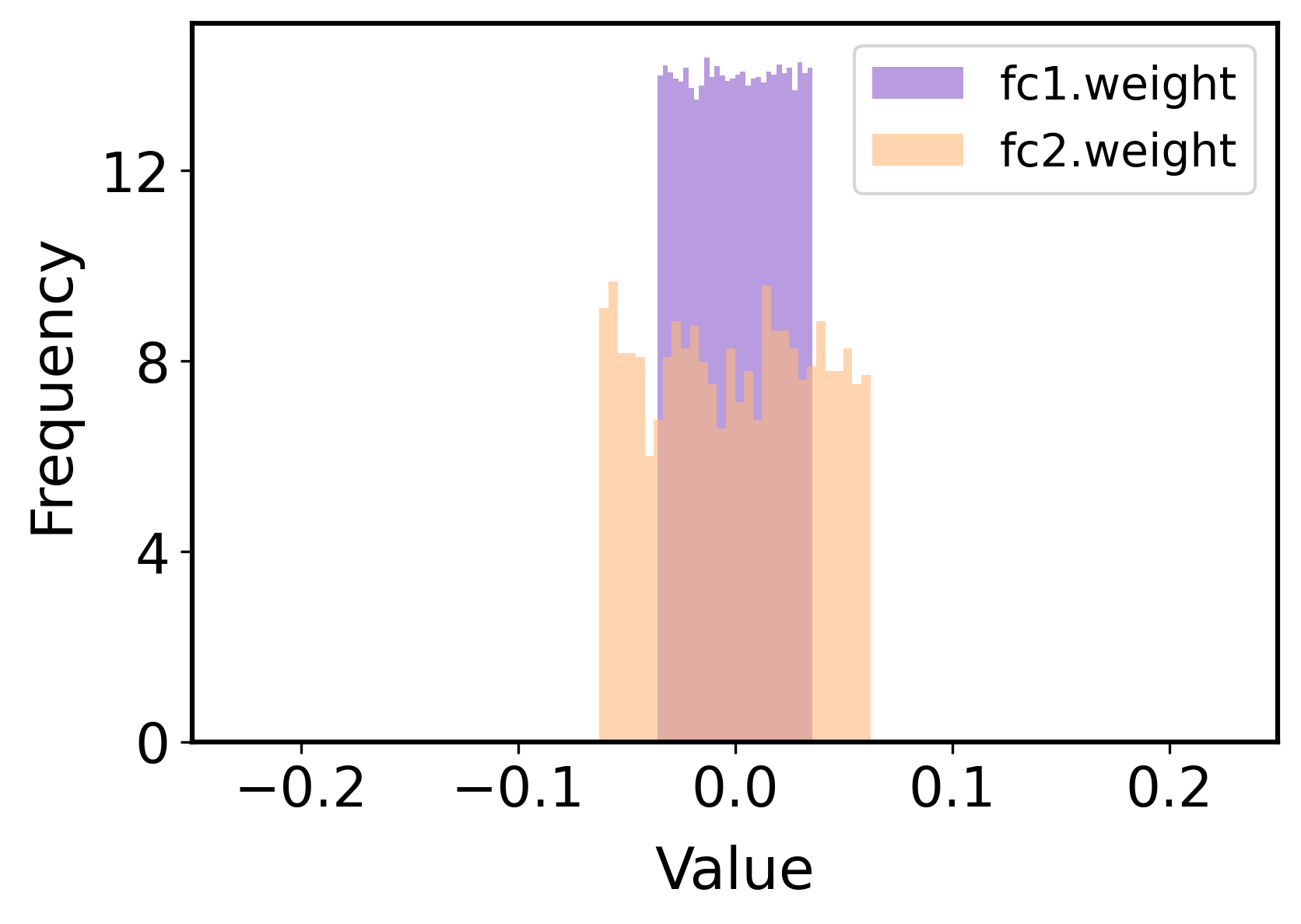}}
    \hfill
    \subfigure[Round 15]{\includegraphics[width=0.24\linewidth]{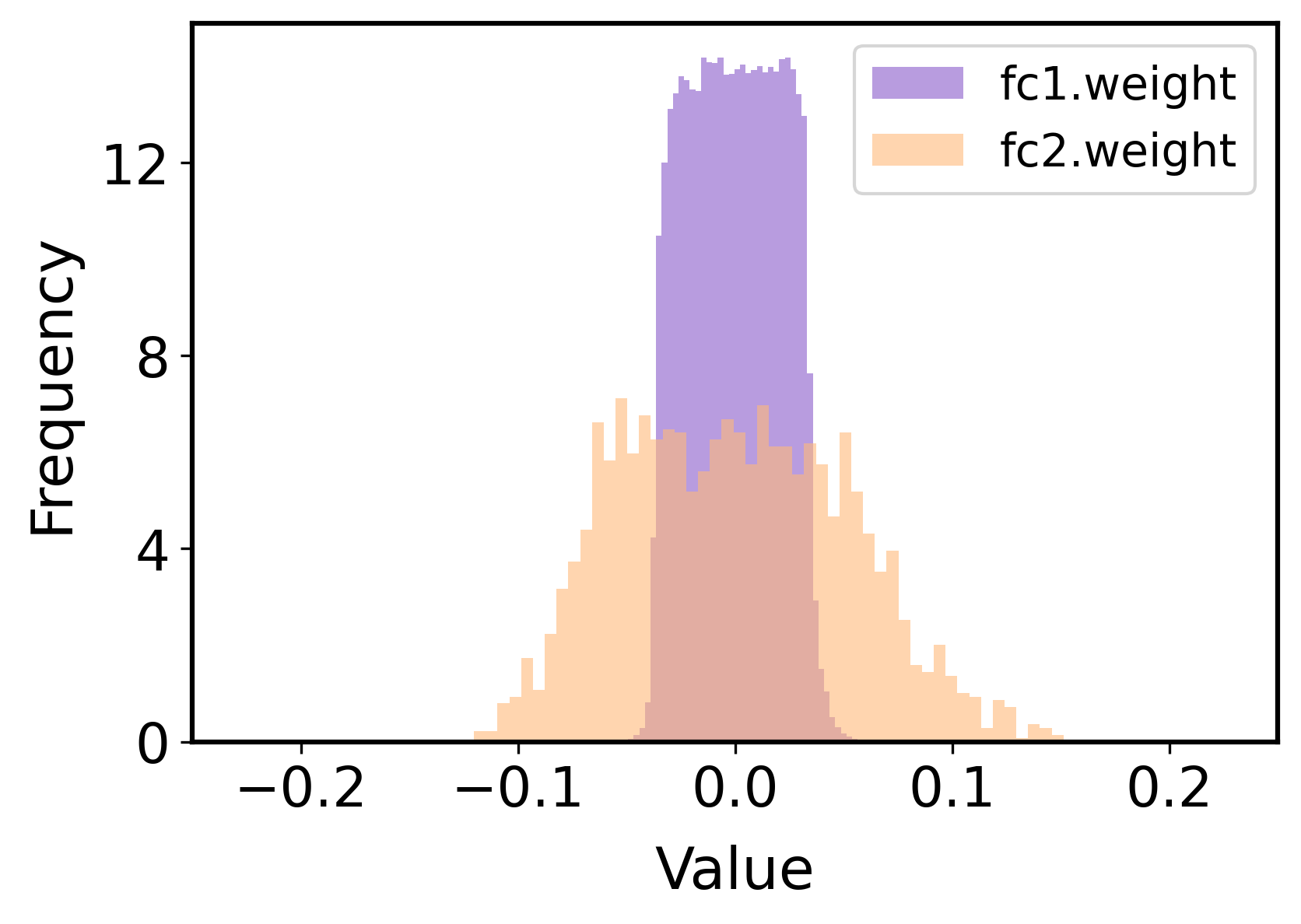}}
    \hfill
    \subfigure[Round 99]{\includegraphics[width=0.24\linewidth]{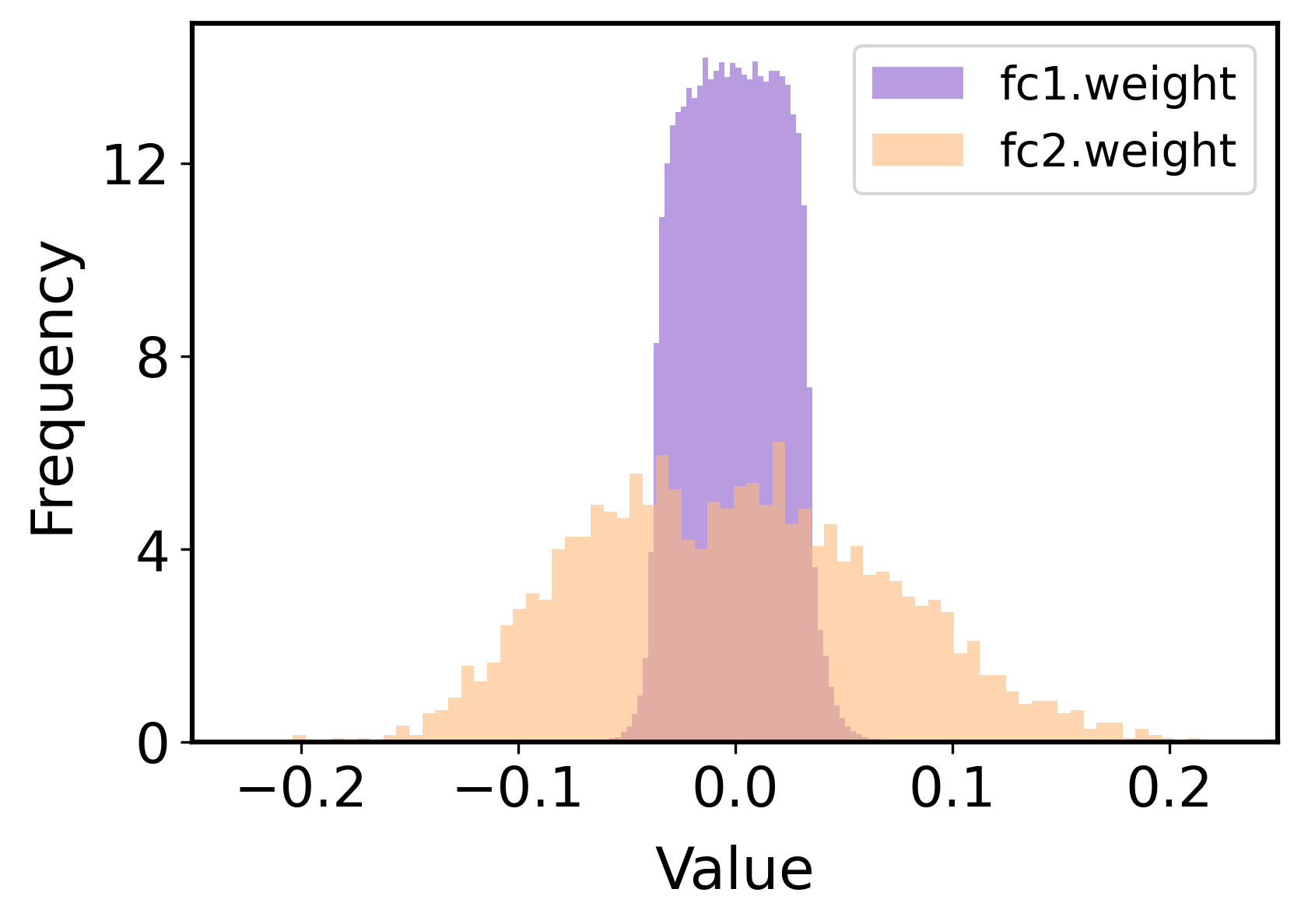}}
    \hfill
    \subfigure[Accuracy vs Rounds]{\includegraphics[width=0.24\linewidth]{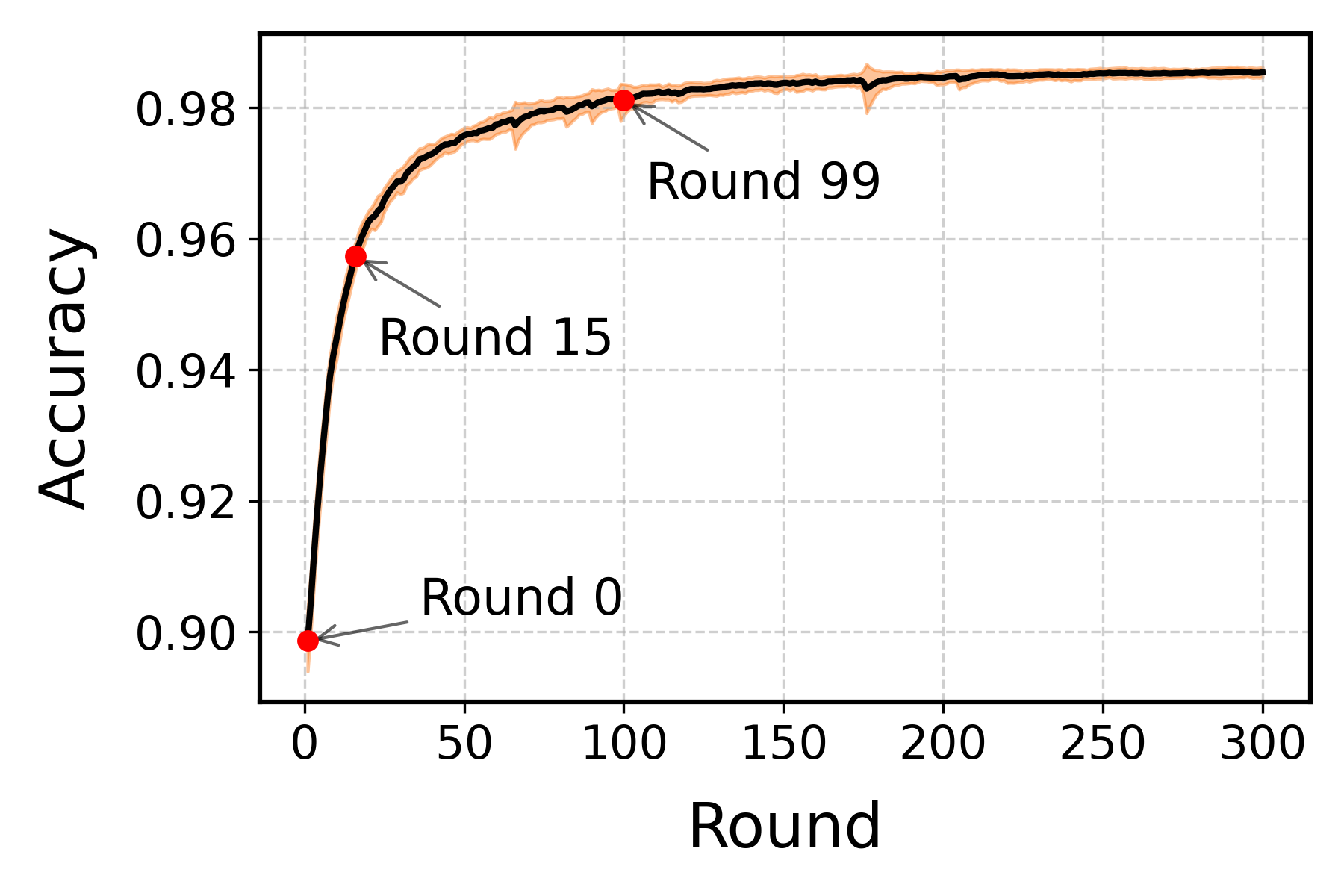}}
    \caption{Layer-wise parameter distribution evolution in a two-layer DNN trained with standard FedAvg. Subfigures (a)--(c) show the distributions of \texttt{fc1.weight} and \texttt{fc2.weight} at rounds 0, 15, and 99. Subfigure (d) shows the test accuracy over communication rounds, with the three selected rounds marked to indicate the initial, intermediate, and converged stages.}
    \label{fig:motivation}
\end{figure*}

Specifically, existing one-bit methods usually apply a uniform binary compression rule across the entire model. Although this design simplifies communication and aggregation, it ignores two important characteristics of deep neural networks. First, parameter distributions can vary substantially across layers. As shown in Fig.~\ref{fig:motivation}, even under standard FedAvg training, the shallow layer (\texttt{fc1.weight}) is more concentrated, whereas the deeper layer (\texttt{fc2.weight}) spreads over a much wider range, and both distributions evolve during training. This suggests that using the same compression rule for all layers is inherently mismatched to their heterogeneous statistics. Second, a single threshold is often too coarse to characterize a layer's distribution, especially when the distribution is broad or changes over time. In contrast, multiple thresholds can partition the sketched parameter space into several intervals, providing broader coverage and a finer representation of layer-wise parameter variations. Therefore, a layer-wise multi-threshold design is better suited to deep models, as it can simultaneously adapt to inter-layer heterogeneity and improve intra-layer quantization resolution, thereby reducing information loss under heterogeneous data distributions.

\begin{figure*}[htb]
	\centering
	\includegraphics[width=0.88\linewidth]{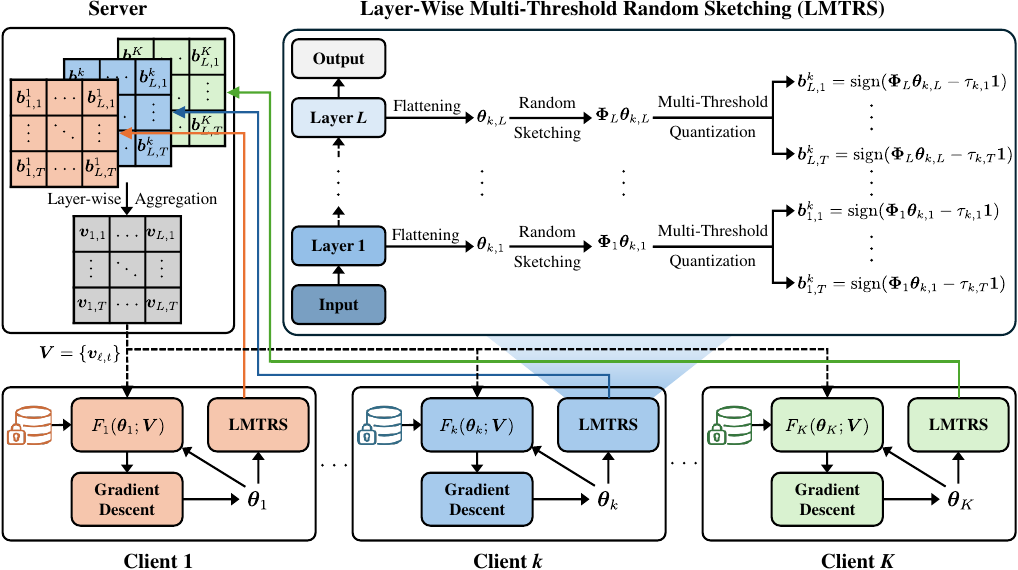}
	\caption{System diagram of the proposed algorithm pFedLMS.}
    \label{fig:SystemDiagram}
\end{figure*}

To address this limitation, we propose a communication-efficient PFL framework named pFedLMS based on layer-wise multi-threshold random sketching (LMTRS). Instead of using a layer-agnostic binary compression rule, LMTRS assigns an independent set of quantization thresholds to each network layer, enabling structure-aware and fine-grained quantization of sketched representations. By integrating LMTRS into both uplink and downlink communication, the proposed framework achieves bidirectional low-bit compression while preserving critical information specific to each layer. The server aggregates compact quantized sketches and broadcasts a concise global consensus signal, which guides local personalization without requiring the transmission of full-precision models.

\subsection{Main Contributions}
The proposed framework in Fig.~\ref{fig:SystemDiagram} aims to improve the communication-accuracy tradeoff by replacing single-threshold one-bit sketching with layer-wise multi-threshold random sketching. The main contributions are summarized as follows:
\begin{enumerate}
    \item[1)] We propose a layer-wise multi-threshold random sketching mechanism for personalized federated learning. Unlike existing one-bit schemes that apply a uniform binary rule to the whole model, the proposed method assigns independent threshold sets to different layers, thereby adapting the quantization geometry to layer-wise parameter distributions.
    
    \item[2)] We formulate a weighted disagreement objective for server aggregation and a multi-threshold consensus regularizer for personalized model training. The server objective admits a closed form solution through weighted majority voting, while the client regularizer aligns local models with the low bit consensus and provides an interval consistency interpretation.
    
    \item[3)] We develop a compact bidirectional communication protocol for the proposed multi-threshold sketches. By encoding the threshold-wise comparison results into a \((T+1)\)-ary representation, the protocol supports both uplink and downlink transmission using only low-bit sketched messages.
    
    \item[4)] We provide theoretical and empirical evidence for the proposed framework. The convergence analysis characterizes the effect of stochastic local updates and cross-round drift caused by changes in the consensus, thresholds, and sketching operators, while empirical results demonstrate that the proposed method improves the balance between communication cost and accuracy compared to representative baselines.
\end{enumerate}

\subsection{Related Work}

PFL addresses data heterogeneity by learning personalized models instead of a single global model \cite{Li2025Dynamic,ZHANG2026104269,zhuang2026personalized,zhang2026FL}.
Existing approaches can be broadly categorized into three classes. Regularization-based methods, such as pFedMe~\cite{dinh2020pfedme} and Ditto~\cite{li2021ditto}, introduce proximal or bilevel formulations to balance global consistency and local adaptation.
Meta-learning-based methods, such as Per-FedAvg~\cite{fallah2020personalized}, aim to learn a shared initialization that enables fast adaptation to each client.
Another line of work focuses on representation splitting, where a shared representation is learned across clients while personalization is achieved through local components, as in FedPer~\cite{arivazhagan2019federated} and FedRep~\cite{collins2021fedrep}.
While these methods improve personalization performance, they typically inherit the communication overhead of standard federated learning, as they rely on transmitting high-dimensional full-precision model parameters or updates.

To reduce communication overhead, recent studies have explored communication-efficient federated learning through quantization, sparsification, and model pruning.
Quantization-based methods reduce the numerical precision of transmitted updates, and recent works further consider rate-constrained formulations that explicitly balance quantization distortion and communication bit rate~\cite{hamidi2025rate}.
Sparsification and dynamic sparse training methods communicate or optimize only a subset of model parameters, reducing both communication and computation costs~\cite{kim2024spafl,li2025pffdst}.
Federated pruning methods further compress client models by removing redundant filters, neurons, or weights during training~\cite{jiang2024fedmp,min2024pf2learning}.
Despite their effectiveness, these methods mainly rely on real-valued compressed updates, sparse structures, or pruned subnetworks, and are not specifically designed for extreme low-bit bidirectional communication in PFL.
Moreover, many of these methods mainly reduce the uplink cost, while the server still needs to transmit high-precision messages or even full model parameters in the downlink.

Another related direction is prototype-based communication.
FedProto~\cite{tan2022fedproto} reduces communication by exchanging compact class-wise prototypes instead of full model parameters.
Although FedProto achieves low communication cost, it mainly performs coordination at the representation level and does not directly compress or align personalized model parameters.
Its performance also depends on the quality of local feature representations and typically requires compatible label spaces and prototype dimensions across clients.
In contrast, our method communicates layer-wise low-bit sketches, enabling direct coordination at the model level while preserving information from individual layers.

More aggressive communication reduction can be achieved by transmitting highly compressed information. A representative example is signSGD~\cite{bernstein2018signsgd}, which communicates the signs of gradients instead of full-precision gradients. This idea has been extended to federated and wireless learning scenarios, including one-bit over-the-air aggregation (OBDA)~\cite{zhu2020obda} and one-bit compressed sensing-based FL (OBCSAA)~\cite{fan2022obcsaa}. Recent studies further improve the stability and convergence of sign-based FL under heterogeneous data. For example, zSignFed~\cite{jin2022zsignfedavg} introduces stochastic sign perturbations into FedAvg-type local training. Another related method, EDEN~\cite{vargaftik2022eden}, improves communication-efficient distributed mean estimation through random rotation, deterministic quantization, and scaling, enabling robust low-bit gradient aggregation under heterogeneous communication budgets and packet losses.

More closely related to our work, pFed1BS~\cite{cheng2026personalized} introduces a PFL framework with bidirectional communication compression via one-bit random sketching. Instead of transmitting full-precision personalized models, pFed1BS uploads one-bit random sketches and broadcasts a one-bit global consensus, thereby achieving extreme bidirectional compression. However, its sketching mechanism still adopts a layer-agnostic one-bit representation, where the same binary quantization rule is applied across the whole model. Such a design does not explicitly account for the layer-wise differences in parameter distributions and quantization sensitivities of deep neural networks. 

Our work bridges PFL and extreme communication compression. Unlike existing PFL methods, we avoid transmitting full-precision personalized models. Unlike existing communication-efficient and one-bit methods, we move beyond the conventional single-threshold design and introduce a layer-wise multi-threshold random sketching framework.
This enables (i) bidirectional low-bit communication, (ii) quantization adapted to layer structures, and (iii) improved representation fidelity under heterogeneous data distributions.

	\section{Problem Formulation}
	
	We consider a PFL system with $K$ clients over $R$ communication rounds. Each client $k \in \{1,\dots,K\}$ owns a local dataset $\mathcal{D}_k=\{\xi_{k,i}\}_{i=1}^{N_k}$, where $N_k=|\mathcal{D}_k|$ denotes the number of local training samples. Client $k$ maintains a personalized model $\bm{\theta}_k \in \mathbb{R}^{n}$, which is decomposed into $L$ layers as
	\begin{equation}
		\bm{\theta}_k
		=
		\big[\bm{\theta}_{k,1}; \dots; \bm{\theta}_{k,L}\big],
		\quad
		\bm{\theta}_{k,\ell} \in \mathbb{R}^{n_\ell},
		\quad
		\sum_{\ell=1}^L n_\ell = n .
	\end{equation}
	Each client aims to minimize its local expected loss
	\begin{equation}
		f_k(\bm{\theta}_k)
		=
		\mathbb{E}_{\xi_k \sim \mathcal{P}_k}
		\big[
		\hat f_k(\bm{\theta}_k;\xi_k)
		\big],
	\end{equation}
	where $\mathcal{P}_k$ denotes the local data distribution, $\xi_k$ denotes a random data sample drawn from $\mathcal{P}_k$, and $\hat f_k(\bm{\theta}_k;\xi_k)$ denotes the sample-wise loss evaluated at $\bm{\theta}_k$ on sample $\xi_k$. The server aggregates client information with weights $\{p_k\}_{k=1}^K$ determined by the local dataset sizes, defined as
	\begin{equation}
		p_k = \frac{N_k}{N},
		\quad
		N = \sum_{k=1}^K N_k ,
	\end{equation}
	which satisfy $p_k \ge 0$ and $\sum_{k=1}^K p_k = 1$.
	
	To enable communication-efficient coordination, we introduce a layer-wise random sketching mechanism. For each layer $\ell$, let $\bm{\Phi}_\ell \in \mathbb{R}^{m_\ell \times n_\ell}$ be a sketching operator with $m_\ell \ll n_\ell$, and consider the sketched representation $\bm{\Phi}_\ell \bm{\theta}_{k,\ell} \in \mathbb{R}^{m_\ell}$. Each layer is further equipped with an ordered set of quantization thresholds
	\begin{equation}
		\boldsymbol{\tau}_\ell = \{ \tau_{\ell,1} < \cdots < \tau_{\ell,T} \}, \quad \ell = 1,\ldots,L.
	\end{equation}
	For each threshold $t$, client $k$ computes one-bit comparisons
	\begin{equation}
		\bm{b}_{\ell,t}^{k} = \mbox{sign}\big( \bm{\Phi}_\ell \bm{\theta}_{k,\ell} - \tau_{\ell,t} \bm{1} \big) \in \{\pm 1\}^{m_\ell},
	\end{equation}
	where $\mathbf{1}$ denotes the all-one vector, $\mbox{sign}(x)=2\mathbb{I}\{x\ge 0\}-1$, i.e., $\mbox{sign}(x)=+1$ for $x\ge 0$ and $\mbox{sign}(x)=-1$ otherwise, and $\mathbb{I}(\cdot)$ denotes the indicator function. The collection of threshold-wise comparisons forms a multi-threshold sketch $\bm{B}_{k,\ell} = (\bm{b}_{\ell,1}^{k}, \dots, \bm{b}_{\ell,T}^{k})$.
	
	The server maintains a layer-wise multi-threshold consensus variable $\bm{V} = \{\bm{v}_{\ell,t}\}_{\ell,t}$ with $\bm{v}_{\ell,t} \in \{\pm 1\}^{m_\ell}$, and updates it via weighted majority voting
	\begin{equation}
		\label{eq:v_lt} \bm{v}_{\ell,t} = \mbox{sign} \Big( \sum_{k=1}^K p_k \, \bm{b}_{\ell,t}^{k} \Big), \quad \forall \ell, t.
	\end{equation}
	The consensus $\bm{V}$ is then broadcast to all clients.
	
	To enforce consistency between local models and the global consensus, we introduce a layer-wise multi-threshold alignment regularizer
	\begin{equation}
		\label{eq:reg} \mathcal{R}(\bm{\theta}_{k}; \bm{V}) = \frac{2}{T}\sum_{\ell=1}^L \sum_{t=1}^T \Big\| \big[\bm{v}_{\ell,t} \odot (\bm{\Phi}_\ell \bm{\theta}_{k,\ell}- \tau_{\ell,t} \bm{1}) \big]_{-} \Big\|_1,
	\end{equation}
	where $[x]_- = \min(x,0)$ is applied element-wise and $\odot$ denotes the element-wise product. Define 
	\begin{equation}
		\mathcal V \triangleq \big{\{} \{\bm v_{\ell,t}\}_{\ell,t} \big| \bm v_{\ell,t}\in\{\pm1\}^{m_\ell} \big{\}}.
	\end{equation}

	Define the server aggregation objective as
	\begin{equation}
		\mathcal J\left(\bm V;\{\bm B_k\}_{k=1}^{K}\right)
		\triangleq
		\frac{1}{T}\sum_{\ell=1}^{L}\sum_{t=1}^{T}\sum_{k=1}^{K}
		p_k\left\|\left[\bm v_{\ell,t}\odot\bm b_{\ell,t}^{k}\right]_{-}\right\|_1,
		\label{eq:server_objective}
	\end{equation}
	where $\bm B_k=\{\bm b_{\ell,t}^{k}\}_{\ell,t}$ denotes the multi-threshold sketch uploaded by client $k$. Since $\bm v_{\ell,t}$ and $\bm b_{\ell,t}^{k}$ are binary vectors, each entry of $\|[\bm v_{\ell,t}\odot\bm b_{\ell,t}^{k}]_{-}\|_1$ equals one when the corresponding signs disagree and zero otherwise. Thus, $\mathcal J$ measures the weighted disagreement between the server consensus and the client sketches.
	
	The server updates the consensus by solving
	\begin{equation}
		\min_{\bm V\in\mathcal V}
		\mathcal J\left(\bm V;\{\bm B_k\}_{k=1}^{K}\right).
		\label{eq:server_aggregation}
	\end{equation}
	
	Lemma~\ref{lem:majority_voting} shows that the majority voting rule in \eqref{eq:v_lt} is a closed form solution of \eqref{eq:server_aggregation}.
	
	\begin{lemma}[Optimality of majority voting]
		\label{lem:majority_voting}
		Problem~\eqref{eq:server_aggregation} admits a coordinatewise closed form solution given by
		\begin{equation}
			\bm v_{\ell,t}^{\star}
			=
			\operatorname{sign}\left(\sum_{k=1}^{K}p_k\bm b_{\ell,t}^{k}\right),
			~ \forall \ell,t.
		\end{equation}
	\end{lemma}
	
	The following lemma shows that the proposed multi-threshold alignment regularizer \eqref{eq:reg} admits an interval consistency interpretation: the aggregated multi-threshold signs induce a unique interval for each coordinate. Moreover, increasing $T$ refines this partition and yields a higher-resolution alignment.
	
	\begin{lemma}[Interval consistency]
		\label{lem:interval}
		Fix a coordinate $(\ell,i)$ and let
		\[
		y_k = (\bm{\Phi}_\ell \bm{\theta}_{k,\ell})(i),\quad
		v_{\ell,t}(i)
		=
		\mbox{\rm sign}\!\left(
		\sum_{k=1}^K p_k\,\mbox{\rm sign}(y_k-\tau_{\ell,t})
		\right).
		\]
		Then there exists $s_{\ell,i}\in\{0,1,\dots,T\}$ such that
		\begin{equation}
			\begin{aligned}
				v_{\ell,t}(i)&=+1,\quad t\le s_{\ell,i},\\
				v_{\ell,t}(i)&=-1,\quad t>s_{\ell,i}.
			\end{aligned}
		\end{equation}
		With the convention $\tau_{\ell,0}=-\infty$ and
		$\tau_{\ell,T+1}=+\infty$, define the induced consistency interval as
		\begin{equation}
			\mathcal I_{\ell,i}
			\triangleq
			\left[
			\tau_{\ell,s_{\ell,i}},
			\tau_{\ell,s_{\ell,i}+1}
			\right].
		\end{equation}
		Then the regularizer in \eqref{eq:reg} enforces interval consistency in the sense that the coordinate-wise penalty vanishes if and only if
		\begin{equation}
			(\bm{\Phi}_\ell \bm{\theta}_{k,\ell})(i)\in \mathcal I_{\ell,i}.
		\end{equation}
	\end{lemma}

	\begin{lemma}[Resolution gain]
		\label{lem:resolution_gain}
		Suppose $T\ge 2$. For layer $\ell$, define the maximum internal interval width as
		\begin{equation}
			\Delta_{\ell,T}
			\triangleq
			\max_{1\le t\le T-1}
			\left(
			\tau_{\ell,t+1}-\tau_{\ell,t}
			\right).
		\end{equation}
		For any coordinate $i$ whose induced interval in Lemma~\ref{lem:interval} is an internal interval, i.e.,
		$s_{\ell,i}\in\{1,\ldots,T-1\}$, we have
		\begin{equation}
			|\mathcal I_{\ell,i}|
			\le
			\Delta_{\ell,T}.
		\end{equation}
		Moreover, if the thresholds are selected as quantiles $q_t=t/(T+1)$ of a reference distribution \(f_{\ell}^{\rm ref}\) whose density is lower bounded by
		$f_{\ell,\min}^{\rm ref}>0$ on $[\tau_{\ell,1},\tau_{\ell,T}]$, then
		\begin{equation}
			\Delta_{\ell,T}
			\le
			\frac{1}{(T+1)f_{\ell,\min}^{\rm ref}}.
		\end{equation}
	\end{lemma}
	
	Lemma~\ref{lem:interval} shows that the aggregated multi-threshold signs always induce a valid consistency interval on the whole real line. Lemma~\ref{lem:resolution_gain} further quantifies the finite-resolution benefit within the internal threshold-covered region $[\tau_{\ell,1},\tau_{\ell,T}]$. The two tail intervals $(-\infty,\tau_{\ell,1}]$ and $[\tau_{\ell,T},+\infty)$ are not used to define a finite interval width. In our implementation, the reference distribution is the layer-wise Gaussian distribution determined by the aggregated mean and variance. Under the quantile choice $q_t=t/(T+1)$, their reference probability masses are controlled by the two extreme quantile levels, while the internal intervals become finer as \(T\) increases; when \(f_{\ell,\min}^{\rm ref}\) is bounded away from zero, this refinement is of order \(O(1/T)\).

	Since $\mathcal{R}$ is non-smooth due to the $\ell_1$ norm and the hinge operator $[\cdot]_{-}$, we adopt a Nesterov smoothing technique to obtain a differentiable approximation, enabling efficient gradient-based optimization and facilitating convergence analysis. Using the identity
	\begin{multline}
		\Big\| \big[ \bm{v}_{\ell,t} \odot (\bm{\Phi}_\ell \bm{\theta}_{k,\ell} - \tau_{\ell,t} \bm{1}) \big]_{-} \Big\|_1 \\
		= \frac{1}{2} \Big( \|\bm{\Phi}_\ell \bm{\theta}_{k,\ell} - \tau_{\ell,t} \bm{1}\|_1 - \langle \bm{v}_{\ell,t}, \bm{\Phi}_\ell \bm{\theta}_{k,\ell} - \tau_{\ell,t} \bm{1} \rangle \Big),
	\end{multline}
	and defining the smooth approximation
	\begin{equation}
		\|\bm{z}\|_{1,\rho} = \max_{\|\bm{u}\|_\infty \le 1} \left\{ \langle \bm{u}, \bm{z} \rangle - \frac{\rho}{2} \|\bm{u}\|_2^2 \right\},
	\end{equation}
	we obtain the smoothed regularizer
	\begin{multline}
		\widetilde{\mathcal{R}}_{\rho}(\bm{\theta}_{k}; \bm{V}) = \frac{1}{T} \sum_{\ell=1}^L \sum_{t=1}^T \Big[ \|\bm{\Phi}_\ell \bm{\theta}_{k,\ell} - \tau_{\ell,t} \bm{1}\|_{1,\rho} \\
		- \langle \bm{v}_{\ell,t}, \bm{\Phi}_\ell \bm{\theta}_{k,\ell} - \tau_{\ell,t} \bm{1} \rangle \Big].
	\end{multline}
	The smoothed regularizer $\widetilde{\mathcal{R}}_\rho(\bm\theta_k;\bm V^r)$ can be viewed as a differentiable relaxation of the hard interval-consistency penalty $\mathcal{R}(\bm\theta_k;\bm V)$, with a uniform approximation error of order $O(\rho)$.
	
	Finally, the proposed method can be formulated as an alternating server and client optimization framework:
	\begin{equation}
		\text{Server:}\quad
		\bm V^\star
		\in
		\arg\min_{\bm V\in\mathcal V}
		\mathcal J\left(\bm V;\{\bm B_k\}_{k=1}^{K}\right),
	\end{equation}
	\begin{equation}
		\text{Clients:}\quad
		\bm\theta_k^\star(\bm V)
		\in
		\arg\min_{\bm\theta_k}
		F_k(\bm\theta_k;\bm V),
	\end{equation}
	where
	\begin{equation}
		F_k(\bm\theta_k;\bm V)
		\triangleq
		f_k(\bm\theta_k)
		+\lambda\widetilde{\mathcal R}_{\rho}(\bm\theta_k;\bm V)
		+\frac{\mu}{2}\|\bm\theta_k\|_2^2.
	\end{equation}
	This formulation separates the discrete server aggregation from the smooth client optimization, enabling efficient communication through low bit sketches while preserving personalization.

\section{Algorithm Design}

\subsection{Communication-Efficient Implementation}
\label{subsec:comm_protocol}

To reduce communication overhead, we adopt a compact representation for transmitting multi-threshold sketches, as shown in Fig. \ref{fig:CommProtocol}. Instead of sending $T$ one-bit comparison vectors per layer, each client aggregates them coordinate-wise into a single $(T+1)$-ary vector, where each entry indicates the interval defined by the ordered thresholds. This $(T+1)$-ary vector is then encoded into a binary bitstream via a base-$(T+1)$ representation and transmitted to the server.

Upon reception, the server decodes the bitstream and reconstructs the threshold-wise comparison vectors. Aggregation is performed using the same majority voting rule as in the theoretical formulation \eqref{eq:v_lt}, ensuring that the communication scheme does not alter the optimization objective. The aggregated results are then re-encoded and broadcast to all clients, where they are decoded and used for local updates.

With this design, both uplink and downlink communication require
\begin{equation}
    \lceil m_\ell \log_2 (T+1) \rceil
\end{equation}
bits per layer, which scales logarithmically with the number of thresholds. When $T=1$, the scheme reduces to one-bit sketching. Detailed implementation is provided in the Supplemental Material. Note that this expression counts the dominant sketch payload. The additional layer-wise statistics and thresholds require only $\mathcal{O}(L)$ and $\mathcal{O}(LT)$ scalar transmissions, respectively, which are negligible compared to the sketch payload for large layers.

\vspace{0.4cm} 
\noindent
\begin{minipage}{\linewidth}
    \centering
    \includegraphics[width=0.9\linewidth]{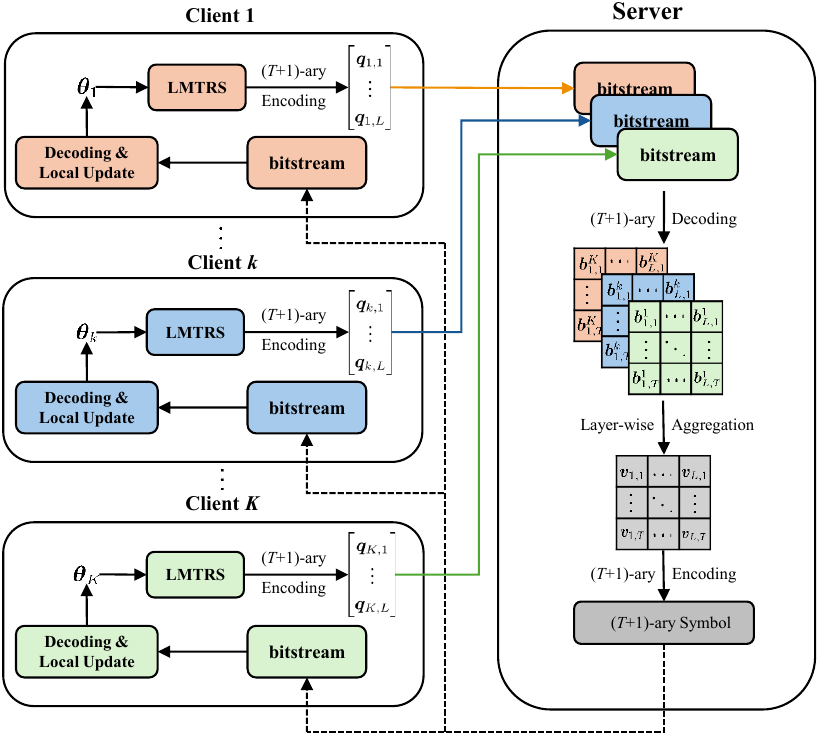}
    \captionof{figure}{Communication protocol for the proposed framework. Here, LMTRS is the block shown in Fig. \ref{fig:SystemDiagram}.}
    \label{fig:CommProtocol}
\end{minipage}
\vspace{0.4cm}

	\subsection{Algorithm}
	The proposed algorithm is presented in Algorithm \ref{alg:pFedLMS}. We index communication rounds by $r=0,1,\ldots,R-1$ and use the round-dependent notation $\bm{\Phi}_\ell^r$, $\boldsymbol{\tau}_\ell^r$, $\bm{B}_{k,\ell}^r$ and $\bm{V}^r$ for the sketching operators, thresholds, threshold-wise comparisons, and consensus. The sketching operators $\{\bm\Phi_\ell^r\}$ are generated from shared random seeds, so they need not be transmitted explicitly.
	
	\textbf{Server update.}
	At each communication round $r$, the server first broadcasts the current layer-wise threshold set $\boldsymbol{\tau}^r=\{\tau_{\ell,t}^r\}_{\ell=1,t=1}^{L,T}$. Each participating client computes a set of one-bit sketches $\{\bm{B}_{k,\ell}^{r}\}_{\ell=1}^L$ by comparing its current layer-wise sketched model with the prescribed threshold levels, i.e.,
	\begin{equation}
		\bm b_{\ell,t}^{k,r}
		=
		\mbox{sign}\!\big(
		\bm\Phi_\ell^{r}\bm\theta_{k,\ell}^{r}
		-
		\tau_{\ell,t}^{r}\bm 1
		\big),
		\quad
		\bm{B}_{k,\ell}^{r}=\{\bm b_{\ell,t}^{k,r}\}_{t=1}^T .
	\end{equation}
	The resulting multi-threshold sketches are sent to the server. The server then updates the consensus by performing weighted majority voting independently across layers and thresholds, i.e.,
	\begin{equation}
		\bm{v}_{\ell,t}^{r}
		=
		\mbox{sign}
		\left(
		\sum_{k=1}^K p_k \bm{b}_{\ell,t}^{k,r}
		\right).
	\end{equation}
	This layer-wise multi-threshold aggregation yields a compact global representation $\bm{V}^r=\{\bm v_{\ell,t}^r\}_{\ell=1,t=1}^{L,T}$, which is broadcast to all clients for the current local update.
	
	\textbf{Client update.}
	Given the broadcast consensus $\bm{V}^r$, each client performs $E$ steps of local stochastic gradient descent to update its personalized model. At each local step, the update takes the form
	\begin{equation}
		\begin{aligned}
			\bm{\theta}_k^{r,e+1} =& \bm{\theta}_k^{r,e} - \eta\Big( \frac{1}{|\mathcal{B}_{k}^{r,e}|} \sum_{\xi_k^{r,e} \in \mathcal{B}_{k}^{r,e}}\nabla \hat f_k(\bm{\theta}_k^{r,e};\xi_k^{r,e}) \\
			&\qquad + \lambda \nabla_{\bm{\theta}_k}\widetilde{\mathcal R}_\rho(\bm{\theta}_k^{r,e};\bm V^r) + \mu \bm{\theta}_k^{r,e} \Big),
		\end{aligned}
	\end{equation}
	where the gradient of $\widetilde{\mathcal R}_\rho$ with respect to the $\ell$-th layer is given by
	\begin{multline}
		\nabla_{\bm{\theta}_{k,\ell}}\widetilde{\mathcal R}_{\rho}(\bm{\theta}_k;\bm{V}^r) \\
		= \frac{1}{T}(\bm{\Phi}_\ell^r)^\top \sum_{t=1}^T \Bigg[ \mathrm{clip}_{[-1,1]} \!\left( \frac{\bm{\Phi}_\ell^r\bm{\theta}_{k,\ell}-\tau_{\ell,t}^r\mathbf 1}{\rho} \right) - \bm{v}_{\ell,t}^r \Bigg] 
		\label{eq:regularizer-gradient}
	\end{multline}
	and $\mathrm{clip}_{[-1,1]}(x)=\max\{-1,\min\{1,x\}\}$.
	
	After $E$ local updates, the client obtains $\bm{\theta}_k^{r+1}$ and computes layer-wise sketched statistics $m_\ell^{k,r+1}=\mathrm{Mean}(\bm{\Phi}_\ell^{r+1} \bm{\theta}_{k,\ell}^{r+1})$ and $s_\ell^{k,r+1}=\mathrm{Var}(\bm{\Phi}_\ell^{r+1} \bm{\theta}_{k,\ell}^{r+1})$, where $\mathrm{Mean}(\cdot)$ and $\mathrm{Var}(\cdot)$ denote the empirical mean and variance computed over all entries of the corresponding layer. These statistics are sent to the server for the next threshold update.
	
	\textbf{Threshold update.}
	At each communication round $r$, after obtaining the layer-wise statistics $\{m_{\ell}^{k,r+1}, s_{\ell}^{k,r+1}\}_{\ell=1}^L$ from the clients, the server aggregates them to obtain the global mean and variance for each layer:
	\begin{equation}
		\begin{aligned}
			\bar\mu_\ell^{r+1} &= \sum_{k=1}^K p_k\, m_\ell^{k,r+1}, \\
			(\bar\sigma_\ell^{r+1})^2 &= \sum_{k=1}^K p_k \Big(s_\ell^{k,r+1} + (m_\ell^{k,r+1}-\bar\mu_\ell^{r+1})^2\Big).
		\end{aligned}
	\end{equation}
	
	Based on the aggregated statistics, the server constructs the threshold set
	$\boldsymbol{\tau}_\ell^{r+1}=\{\tau_{\ell,t}^{r+1}\}_{t=1}^T$ via
	\begin{align}
		\tau_{\ell,t}^{r+1} = \bar\mu_\ell^{r+1} + \bar\sigma_\ell^{r+1}\,\Upsilon^{-1}(q_t), \quad
		q_t = \frac{t}{T+1},
	\end{align}
	for $t=1,\ldots,T$, where $\Upsilon^{-1}(\cdot)$ denotes the inverse cumulative distribution function (CDF) of the standard normal distribution. The resulting thresholds are then broadcast to all clients and used for multi-threshold sketching in the next round.
	
	\begin{algorithm}
		\caption{pFedLMS: Personalized Federated Learning via Layer-Wise Multi-Threshold Random Sketching}
		\label{alg:pFedLMS}
		\begin{algorithmic}[1]
			\STATE {\bfseries Input:} $R,E,\eta,\lambda,\mu,\rho,\{\bm{\Phi}_\ell^r\}, \bm w^0$
			
			\STATE {\bfseries Client initialization:} Each client is initialized with the shared model $\bm w^0$ and performs $E$ steps of local updates to obtain its initial personalized model $\bm{\theta}_k^0$, for $k=1,\dots,K$.
			
			\STATE $m_\ell^{k,0} = \mathrm{Mean}(\bm{\Phi}_\ell^0 \bm{\theta}_{k,\ell}^0), 
			s_\ell^{k,0} = \mathrm{Var}(\bm{\Phi}_\ell^0 \bm{\theta}_{k,\ell}^0),~ \forall k,\ell$
			
			\FOR{$r = 0$ to $R-1$}
			
			\STATE $\{\bm\tau_{\ell,t}^{r}\}
			\leftarrow
			\texttt{ThresholdUpdate}(\{m_\ell^{k,r}, s_\ell^{k,r}\})$
			
			\STATE Broadcast $\{\bm\tau_{\ell,t}^{r}\}$ \hfill // send thresholds
			
			\FOR{\textbf{each} client $k$ \textbf{in parallel}}
			\FOR{each $(\ell,t)$}
			\STATE $\bm b_{\ell,t}^{k,r}
			=
			\mbox{sign}(\bm\Phi_\ell^{r}\bm\theta_{k,\ell}^{r}
			- \bm\tau_{\ell,t}^{r}\bm 1)$
			\hfill // sketching
			\ENDFOR
			\ENDFOR
			
			\FOR{each $(\ell,t)$}
			\STATE $\bm v_{\ell,t}^{r}
			\leftarrow
			\mbox{sign}(\sum_{k=1}^K p_k \bm b_{\ell,t}^{k,r})$
			\hfill // majority voting
			\ENDFOR

			\STATE Broadcast $\bm V^r$ \hfill // global consensus
			
			\FOR{\textbf{each} client $k$ \textbf{in parallel}}
			\STATE $\bm\theta_k^{r+1}, \{m_\ell^{k,r+1}\}, \{s_\ell^{k,r+1}\}
			\leftarrow
			\texttt{ClientUpdate}(k,r,$ $\bm\theta_k^{r},\bm V^r)$
			\ENDFOR
			\ENDFOR
			
			\hdashrule{\linewidth}{0.4pt}{2pt 2pt}
			
			\STATE {\bfseries Function} \texttt{ThresholdUpdate}$(\{m_\ell^{k,r}, s_\ell^{k,r}\})$:
			
			\FOR{$\ell=1,\dots,L$}
			\STATE $\bar\mu_\ell^{r} = \sum_{k=1}^K p_k m_\ell^{k,r}$ \hfill // global mean
			
			\STATE $\bar\sigma_\ell^{r}
			=
			\sqrt{\sum_{k=1}^K p_k(s_\ell^{k,r} + (m_\ell^{k,r}-\bar\mu_\ell^{r})^2)}$
			\hfill // global std
			
			\FOR{$t=1,\dots,T$}
			\STATE $q_t = \frac{t}{T+1}$ \hfill // quantile level
			
			\STATE $\bm\tau_{\ell,t}^{r}
			=
			\bar\mu_\ell^{r}
			+
			\bar\sigma_\ell^{r} \Upsilon^{-1}(q_t)$
			\hfill // threshold
			\ENDFOR
			\ENDFOR
			
			\STATE {\bfseries return} $\{\bm\tau_{\ell,t}^{r}\}$
			
			\hdashrule{\linewidth}{0.4pt}{2pt 2pt}
			
			\STATE {\bfseries Function} \texttt{ClientUpdate}$(k,r,\bm\theta_k^{r},\bm V^r)$:
			
			\STATE $\bm\theta_k^{r,0} \leftarrow \bm\theta_k^{r}$
			
			\FOR{$e = 0$ to $E-1$}
			\STATE Sample $\mathcal B_k^{r,e}$ and compute
			$\widehat{\bm g}_k^{r,e}
			=\widehat{\bm g}_k(\bm\theta_k^{r,e};\mathcal B_k^{r,e})$. \\
			\hfill // SGD step
			
			\FOR{$\ell = 1,\dots,L$}
			\STATE $\bm h_{k,\ell}^{r,e}
			=
			\nabla_{\bm\theta_{k,\ell}}
			\widetilde {\mathcal R}_{\rho}(\bm\theta_k^{r,e};\bm V^r)$
			\hfill // alignment regularizer
			\ENDFOR
			
			\STATE $\bm h_k^{r,e} = [\bm h_{k,1}^{r,e};\dots;\bm h_{k,L}^{r,e}]$
			
			\STATE $\bm\theta_k^{r,e+1} =\bm\theta_k^{r,e} -\eta\left(
			\widehat{\bm g}_k^{r,e} +\lambda\bm h_k^{r,e} +\mu\bm\theta_k^{r,e}
			\right)$
			\hfill // local update
			\ENDFOR
			
			\STATE $\bm\theta_k^{r+1} \leftarrow \bm\theta_k^{r,E}$
			
			\FOR{$\ell = 1,\dots,L$}
			\STATE $m_\ell^{k,r+1} = \mathrm{Mean}(\bm{\Phi}_\ell^{r+1} \bm\theta_{k,\ell}^{r+1})$
			\STATE $s_\ell^{k,r+1} = \mathrm{Var}(\bm{\Phi}_\ell^{r+1} \bm\theta_{k,\ell}^{r+1})$
			\ENDFOR
			
			\STATE {\bfseries return} $\bm\theta_k^{r+1}, \{m_\ell^{k,r+1}\}, \{s_\ell^{k,r+1}\}$
			
		\end{algorithmic} 
	\end{algorithm}
	
	\section{Theoretical Analysis}
	
	In this section, we establish a bound on the average stationarity measure of the client objectives across communication rounds. Compared with standard personalized federated optimization, the key additional difficulty is that the server-side consensus signal is updated round by round through the threshold-wise aggregation rule
	$\bm v_{\ell,t}^{r+1}=\mbox{sign} \left(\sum_{k=1}^K p_k \bm b_{\ell,t}^{k,r+1}\right)$,
	so the client objective is time-varying and depends explicitly on both the sketch comparisons and the threshold sequence. For fixed $\bm{V}^r=\{\bm{v}_{\ell,t}^r\}_{\ell,t}$, define the round-$r$ client objective by
	\begin{equation}
		F_k(\bm{\theta};\bm{V}^r) \triangleq f_k(\bm{\theta}) + \lambda \widetilde{\mathcal R}_{\rho}(\bm{\theta};\bm{V}^r) + \frac{\mu}{2}\|\bm{\theta}\|_2^2, 
		\label{eq:local-objective}
	\end{equation}
	where
	\begin{multline}
		\widetilde{\mathcal R}_{\rho}(\bm{\theta};\bm{V}^r) = \frac{1}{T} \sum_{\ell=1}^L \sum_{t=1}^T \Big( \|\bm{\Phi}_\ell^r \bm{\theta}_\ell-\tau_{\ell,t}^r\mathbf 1\|_{1,\rho} \\
		- \langle \bm{v}_{\ell,t}^r,\, \bm{\Phi}_\ell^r \bm{\theta}_\ell-\tau_{\ell,t}^r\mathbf 1 \rangle \Big). 
		\label{eq:regularizer}
	\end{multline}
	For simplicity, the dependence of $F_k(\cdot;\bm V^r)$ on $\{\bm\Phi_\ell^r\}$ and $\{\tau_{\ell,t}^r\}$ is left implicit.
	
	We first state the assumptions used in the analysis. Assumptions~\ref{ass:smooth}--\ref{ass:lower} are standard in nonconvex stochastic federated optimization. The remaining assumptions are tailored to the proposed framework, which combines random sketching with multiple thresholds. In particular, Assumption~\ref{ass:sketch} controls the stability of random projections, while Assumption~\ref{ass:threshold} ensures that the threshold sequence is properly defined and uniformly bounded.

	\begin{assumption}
		\label{ass:smooth}
		For each client $k\in\{1,\ldots,K\}$, the local objective $f_k$ is continuously differentiable and $L_f$-smooth, i.e.,
		\begin{equation}
			\|\nabla f_k(\bm{\theta})-\nabla f_k(\bm{\theta}')\|_2 \le L_f \|\bm{\theta}-\bm{\theta}'\|_2, ~ \forall \bm{\theta},\bm{\theta}'\in\mathbb{R}^n. 
			\label{eq:smoothness}
		\end{equation}
	\end{assumption}

	\begin{assumption}
		\label{ass:sgd}
		For each client $k$ and any model parameter $\bm{\theta}$, define the mini-batch stochastic gradient as
		\begin{equation}
			\widehat{\bm g}_{k}(\bm{\theta};\mathcal B)
			\triangleq
			\frac{1}{|\mathcal B|}
			\sum_{\xi\in \mathcal B}
			\nabla \hat f_k(\bm{\theta};\xi).
			\label{eq:g_hat}
		\end{equation}
		It satisfies
		\begin{equation}
			\mathbb{E}\!\left[
			\widehat{\bm g}_{k}(\bm{\theta};\mathcal B) \mid \bm{\theta}
			\right]
			=
			\nabla f_k(\bm{\theta}),
		\end{equation}
		and there exists a constant $\sigma^2>0$ such that
		\begin{equation}
			\mathbb{E}\!\left[
			\left\|
			\widehat{\bm g}_{k}(\bm{\theta};\mathcal B)
			-
			\nabla f_k(\bm{\theta})
			\right\|_2^2 \mid \bm{\theta}
			\right]
			\le
			\sigma^2,
			\quad \forall \bm{\theta}, k.
		\end{equation}
	\end{assumption}
	
	\begin{assumption}
		\label{ass:task-grad}
		For each client $k$ and any model parameter $\bm{\theta}$, the mini-batch stochastic gradient in \eqref{eq:g_hat} has bounded second moment. That is, there exists a constant $G>0$ such that
		\begin{equation}
			\mathbb{E}\!\left[
			\|\widehat{\bm g}_{k}(\bm{\theta};\mathcal B)\|_2^2
			\right]
			\le G^2,
			\quad \forall \bm{\theta}, k.
			\label{eq:task-grad-bound}
		\end{equation}
	\end{assumption}

	\begin{assumption}
		\label{ass:lower}
		Define the round-$r$ global potential as 
		\begin{equation}
			\Psi^r \triangleq \sum_{k=1}^K p_k F_k(\bm{\theta}_k^r;\bm{V}^r),
		\end{equation}
		where $F_k$ is the local regularized objective defined in \eqref{eq:local-objective}. Assume that there exists $\Psi_{\inf}>-\infty$ such that $\Psi^r \ge \Psi_{\inf},\forall r$.
	\end{assumption}
	
	\begin{assumption}
		\label{ass:sketch}
		For each layer $\ell$ and round $r$, the sketching operator $\bm{\Phi}_\ell^r\in\mathbb{R}^{m_\ell\times n_\ell}$ satisfies
		$\|\bm{\Phi}_\ell^r\|_2 \le \kappa_\ell,$ where $\kappa_\ell>0$ is a known constant.
	\end{assumption}
	
	\begin{remark}
		Assumption~\ref{ass:sketch} imposes a boundedness condition on the random sketching operator. It prevents the sketching map from amplifying model perturbations without control. This condition holds for commonly used normalized sketching operators, including structured projections based on the Hadamard transform as used in pFed1BS. It also holds with high probability for normalized Gaussian random projections and normalized random projections with sub-Gaussian entries.
	\end{remark}

	\begin{assumption}
		\label{ass:threshold}
		For each layer $\ell$ and round $r$, the threshold set $\boldsymbol{\tau}_\ell^r=\{\tau_{\ell,1}^r<\cdots<\tau_{\ell,T}^r\}$ is ordered and bounded.
		There exists $\tau_{\max}>0$ such that $|\tau_{\ell,t}^r|\le \tau_{\max}, \forall \ell,t,r.$
	\end{assumption}
	
	\begin{remark}
		Assumption~\ref{ass:threshold} guarantees that the threshold comparisons are properly defined and that the regularizer terms involving thresholds remain uniformly controlled. For any fixed number of thresholds \(T\), the Gaussian quantile thresholds used in our algorithm are finite. If a uniform bound independent of \(T\) is desired, clipped quantile levels can be used.
	\end{remark}
	
	%
	

	\begin{lemma}[Bounded model second moment]
		\label{lem:bounded-model}
		Suppose Assumptions~\ref{ass:task-grad}, and \ref{ass:sketch} hold. If the learning rate satisfies $0<\eta \le 1/\mu,$
		then the client iterates satisfy
		$\mathbb{E}\!\left[\|\bm{\theta}_k^{r,e}\|_2^2\right]
		\le B_{\theta}^2, \forall k,r,e,$
		where
		\begin{equation*}
			B_{\theta} \triangleq \max\left\{ \sqrt{\max_{1\le k\le K} \mathbb{E} \|\bm{\theta}_k^0\|_2^2},\, \sqrt{\frac{C_B}{\eta\mu}} \right\}
		\end{equation*}
		and
		\begin{equation*}
			C_B \triangleq \frac{2\eta}{\mu} \left( G^2+4\lambda^2 \sum_{\ell=1}^L \kappa_\ell^2 m_\ell \right).
		\end{equation*}
	\end{lemma}

	\begin{lemma}[Smoothness of the objective]
		\label{lem:F-smooth}
		Under Assumptions~\ref{ass:smooth} and~\ref{ass:sketch}, for any fixed round $r$, the function $F_k(\cdot;\bm{V}^r)$ is $L_F$-smooth, where 
		$L_F = L_{f} + \max_{\ell}\frac{\lambda \kappa_\ell^2}{\rho} + \mu.$

	\end{lemma}
	
	\begin{lemma}[One-step descent]
		\label{lem:one-step}
		Suppose Assumptions~\ref{ass:smooth}--\ref{ass:sketch} hold.
		If $\eta \le 1/L_F$, then for every client $k$, round $r$, and local step $e$,
		\begin{multline}
			\mathbb{E}\!\left[ F_k(\bm{\theta}_{k}^{r,e+1};\bm{V}^r) \,\middle|\, \bm{\theta}_{k}^{r,e}\right] \\
			\le\, F_k(\bm{\theta}_{k}^{r,e};\bm{V}^r) -\frac{\eta}{2} \|\nabla F_k(\bm{\theta}_{k}^{r,e};\bm{V}^r)\|_2^2 +\frac{L_F\eta^2}{2}\sigma^2. 
			\label{eq:one-step-descent}
		\end{multline}
	\end{lemma}
	
	Summing \eqref{eq:one-step-descent} over $e=0,\ldots,E-1$ and then over $k$ with weights $\{p_k\}$ yields
	\begin{multline}
		\sum_{k=1}^K p_k\, \mathbb{E}\!\left[F_k(\bm{\theta}_k^{r+1};\bm{V}^r)\right] \le \sum_{k=1}^K p_k\, \mathbb{E}\!\left[F_k(\bm{\theta}_k^{r};\bm{V}^r)\right] \\
		-\frac{\eta}{2} \sum_{k=1}^K p_k \sum_{e=0}^{E-1} \mathbb{E}\|\nabla F_k(\bm{\theta}_k^{r,e};\bm{V}^r)\|_2^2 +\frac{E L_F\eta^2}{2}\sigma^2. 
		\label{eq:round-descent}
	\end{multline}
	
	The following lemma provides a conservative bound on the cross-round objective drift caused by changes in the consensus, thresholds, and sketching operators.
	
	\begin{lemma}[Cross-round objective drift] \label{lem:threshold_update_aware_drift} Suppose Assumptions~\ref{ass:sketch} and~\ref{ass:threshold} hold, and let $B_\theta$ be the second-moment bound in Lemma~\ref{lem:bounded-model}. Define the sketch refresh indicator as $\chi_\ell^r \triangleq \mathbb I\{\bm{\Phi}_\ell^{r+1}\neq\bm{\Phi}_\ell^r\}.$
		For each layer $\ell$, define 
		\begin{equation} 
			\beta_\ell^r \triangleq 4\sqrt{m_\ell}\kappa_\ell B_\theta\chi_\ell^r + 2\sqrt{m_\ell}\kappa_\ell B_\theta + 6m_\ell\tau_{\max},
		\end{equation}
		and let $\beta^r\triangleq\sum_{\ell=1}^L\beta_\ell^r.$ Then, for any client $k$ and round $r$, 
		\begin{equation} 
			\mathbb E\!\left[ \left| F_k(\bm{\theta}_k^{r+1};\bm V^{r+1}) - F_k(\bm{\theta}_k^{r+1};\bm V^r) \right| \right] \le \lambda\beta^r. \label{eq:coarse-drift-bound}
		\end{equation} 
	\end{lemma}
	
	We now present the main convergence result. 
	
	\begin{theorem} \label{thm:convergence} 
		Suppose Assumptions~\ref{ass:smooth}--\ref{ass:threshold} hold. If $0 < \eta \le {1}/{L_F}$, then the iterates generated by pFedLMS satisfy 
		\begin{align} 
			&\frac{1}{R} \sum_{r=0}^{R-1} \frac{1}{E} \sum_{e=0}^{E-1} \sum_{k=1}^K p_k \mathbb E\!\left[ \left\| \nabla F_k(\bm{\theta}_k^{r,e};\bm V^r) \right\|_2^2 \right] \nonumber\\ 
			&\le \frac{2(\mathbb{E}[\Psi^0]-\Psi_{\inf})}{\eta ER} + L_F\eta\sigma^2 + \frac{2\lambda}{\eta ER} \sum_{r=0}^{R-1}\beta^r, \label{eq:main-convergence} 
		\end{align} 
		where $\beta^r$ is defined in Lemma~\ref{lem:threshold_update_aware_drift}.
	\end{theorem}

	Theorem~\ref{thm:convergence} bounds the average squared gradient norm of the client objectives over communication rounds. Since the consensus, thresholds, and sketching operators vary with $r$, the result measures average stationarity rather than convergence to a fixed objective. The optimization term decreases at the rate $O(1/(\eta ER))$, while the remaining error is determined by stochastic gradient noise and objective drift across rounds.
	
	The term $L_F\eta\sigma^2$ is caused by stochastic gradient noise and can be reduced by using a smaller learning rate or larger mini-batches. The term involving $\beta^r$ provides a conservative bound on the variation caused by changes in the consensus, thresholds, and sketching operators. The first component of $\beta_\ell^r$ depends explicitly on the sketch refresh indicator $\chi_\ell^r$ and vanishes when the sketching operator for layer $\ell$ is fixed. The remaining components arise from the worst-case bounds on consensus changes and threshold displacements. Therefore, they may overestimate the actual cross-round drift, particularly when the consensus and thresholds vary slowly during training. 
	
	The regularization parameter $\lambda$ controls the strength of alignment between personalized models and the global low-bit consensus. A smaller $\lambda$ reduces the drift term in \eqref{eq:main-convergence}, but may also weaken the benefit of collaborative consensus learning. Moreover, $L_F$ depends on $\lambda$, so the stochastic gradient term is also affected by the regularization strength. Therefore, $\lambda$ should be selected to balance optimization stability, personalization, and consensus alignment.

\section{Experiments}
\subsection{Experimental Setup}
\subsubsection{Datasets}
Experiments are conducted on five datasets: MNIST, FMNIST, SVHN, CIFAR-10, and CIFAR-100. 
Following common practice in federated learning, the data is distributed across 20 clients in a non-IID manner. Specifically, a Dirichlet distribution with $\alpha = 0.1$ and $\alpha = 0.5$ is adopted to induce different levels of label distribution heterogeneity.
	
	\subsubsection{Compared Methods}
	We benchmark the proposed pFedLMS against a diverse set of representative baselines, including both traditional federated learning methods and PFL approaches: FedAvg \cite{mcmahan2017fedavg}, OBDA \cite{zhu2020obda}, OBCSAA \cite{fan2022obcsaa}, FedProto \cite{tan2022fedproto}, EDEN \cite{vargaftik2022eden}, zSignFed \cite{jin2022zsignfedavg}, and pFed1BS \cite{cheng2026personalized}. 
	
	\subsubsection{Implementation Details}
	The proposed framework is implemented in PyTorch and executed on an NVIDIA RTX 5090 GPU. We simulate a federated learning environment with 20 clients, utilizing a DNN for MNIST and FMNIST, VGG8 for CIFAR-10, and VGG16 for CIFAR-100.  The local batch size is set to 64 for all experiments. Full client participation is assumed in each communication round, and the local training epochs and total communication rounds are appropriately scaled within the ranges of [20, 50] and [200, 350]. For all compared methods, the key hyperparameters are carefully tuned according to their recommended settings and further adjusted on the validation set when necessary, so as to ensure a fair and competitive comparison. For the proposed pFedLMS, the sketching operators are refreshed at every communication round. They are generated locally from synchronized random seeds and therefore do not incur additional communication overhead. Following pFed1BS~\cite{cheng2026personalized}, we use randomized Hadamard projection operators for efficient sketching. Unless otherwise specified, we set $T=7$ for pFedLMS in the following experiments, which corresponds to an $8$-level interval representation and requires $3$ bits per sketched coordinate. 
	
	\subsubsection{Evaluation Metrics}
	We use two primary metrics to evaluate the performance of the algorithms:
	\begin{itemize}
		\item \textbf{Maximum Accuracy:} It is defined as the highest average accuracy achieved on a held-out test set. To ensure statistical reliability, this value is calculated by averaging the results across three to five independent experimental runs using different seeds.
		\item \textbf{Communication Cost:} 
		It is quantified as the total number of bits exchanged between the central server and all participating clients during one communication round. Specifically, it is calculated as the sum of the uplink and downlink communication costs for one client, multiplied by the number of participating clients.
		
	\end{itemize}

	\begin{table*}[t]
		\centering
		\footnotesize
		\setlength{\tabcolsep}{10pt}
		\renewcommand {\arraystretch}{1.15}
		\begin{tabular}{l | c | c | c | c | c}
			\toprule
			\begin{tabular}[c]{@{}l@{}}Method \\ ($\alpha=0.5$)\end{tabular} 
			& MNIST (Acc. \%)
			& FMNIST (Acc. \%)
			& SVHN (Acc. \%)
			& CIFAR-10 (Acc. \%)
			& CIFAR-100 (Acc. \%) \\
			\midrule
			FedAvg   & $98.55 \pm 0.08$ & $92.32 \pm 0.81$ & $92.66 \pm 0.54$ & $86.57 \pm 0.34$ & $61.14 \pm 0.58$ \\
			OBDA     & $91.12 \pm 1.21$ & $83.14 \pm 0.54$ & $36.40 \pm 3.19$ & $37.91 \pm 1.97$ & $26.85 \pm 0.26$ \\
			OBCSAA   & $93.81 \pm 0.28$ & $88.54 \pm 0.85$ & $59.28 \pm 2.31$ & $61.04 \pm 1.57$ & $14.19 \pm 2.35$ \\
			FedProto & $96.21 \pm 0.21$ & $91.11 \pm 0.78$ & $89.62 \pm 0.65$ & $79.03 \pm 0.53$ & $44.30 \pm 1.16$ \\
			EDEN     & $98.12 \pm 0.11$ & $89.24 \pm 2.24$ & $85.08 \pm 3.65$ & $79.85 \pm 0.37$ & $38.81 \pm 1.39$ \\
			zSignFed & $94.98 \pm 0.49$ & $86.68 \pm 0.67$ & $41.31 \pm 1.91$ & $50.83 \pm 4.11$ & $41.52 \pm 1.51$ \\
			pFed1BS  & $95.10 \pm 0.33$ & $89.35 \pm 0.81$ & $82.17 \pm 2.57$ & $72.97 \pm 0.36$ & $33.98 \pm 0.52$ \\
			\midrule
			\textbf{pFedLMS} 
			& $\mathbf{97.45 \pm 0.32}$ 
			& $\mathbf{91.57 \pm 0.92}$ 
			& $\mathbf{91.66 \pm 0.25}$ 
			& $\mathbf{82.14 \pm 0.46}$ 
			& $\mathbf{47.47 \pm 0.97}$ \\
			\bottomrule
		\end{tabular}
    \caption{Comparison of maximum test accuracy under the Dirichlet non-IID setting with $\alpha=0.5$.}
	\label{tab:main_accuracy_results_alpha05}
	\end{table*}
	
	\begin{table*}[t]
		\centering
		\footnotesize
		\setlength{\tabcolsep}{10pt}
		\renewcommand{\arraystretch}{1.15}
		\begin{tabular}{l | c | c | c | c | c}
			\toprule
			\begin{tabular}[c]{@{}l@{}}Method \\ ($\alpha=0.1$)\end{tabular} 
			& MNIST (Acc. \%)
			& FMNIST (Acc. \%)
			& SVHN (Acc. \%)
			& CIFAR-10 (Acc. \%)
			& CIFAR-100 (Acc. \%) \\
			\midrule
			FedAvg   & $99.19 \pm 0.15$ & $97.67 \pm 0.79$ & $96.52 \pm 0.32$ & $92.62 \pm 1.21$ & $74.60 \pm 0.99$ \\
			OBDA     & $86.84 \pm 0.95$ & $77.50 \pm 3.96$ & $15.09 \pm 4.45$ & $18.12 \pm 4.32$ & $21.43 \pm 1.48$ \\
			OBCSAA   & $96.35 \pm 0.89$ & $95.90 \pm 1.15$ & $75.09 \pm 5.04$ & $78.76 \pm 4.35$ & $20.35 \pm 3.36$ \\
			FedProto & $98.61 \pm 0.23$ & $97.58 \pm 0.89$ & $95.32 \pm 0.68$ & $91.12 \pm 1.26$ & $66.29 \pm 0.45$ \\
			EDEN     & $98.83 \pm 0.20$ & $96.93 \pm 0.99$ & $93.19 \pm 1.81$ & $88.71 \pm 2.85$ & $58.28 \pm 1.21$ \\
			zSignFed & $95.63 \pm 0.45$ & $82.77 \pm 1.40$ & $40.66 \pm 1.73$ & $27.59 \pm 2.47$ & $30.81 \pm 0.78$ \\
			pFed1BS  & $98.20 \pm 0.31$ & $97.18 \pm 0.94$ & $88.18 \pm 3.51$ & $87.46 \pm 2.11$ & $55.93 \pm 0.44$ \\
			\midrule
			\textbf{pFedLMS} 
			& $\mathbf{98.78 \pm 0.22}$ 
			& $\mathbf{97.63 \pm 0.85}$ 
			& $\mathbf{95.42 \pm 0.31}$ 
			& $\mathbf{91.15 \pm 1.27}$ 
			& $\mathbf{66.78 \pm 0.74}$ \\
			\bottomrule
		\end{tabular}
        \caption{Comparison of maximum test accuracy under the Dirichlet non-IID setting with $\alpha=0.1$.}
	\label{tab:main_accuracy_results_alpha01}
	\end{table*}

\subsection{Experimental Results}
	\begin{figure}[b]
		\centering
		\includegraphics[width=0.8\linewidth]{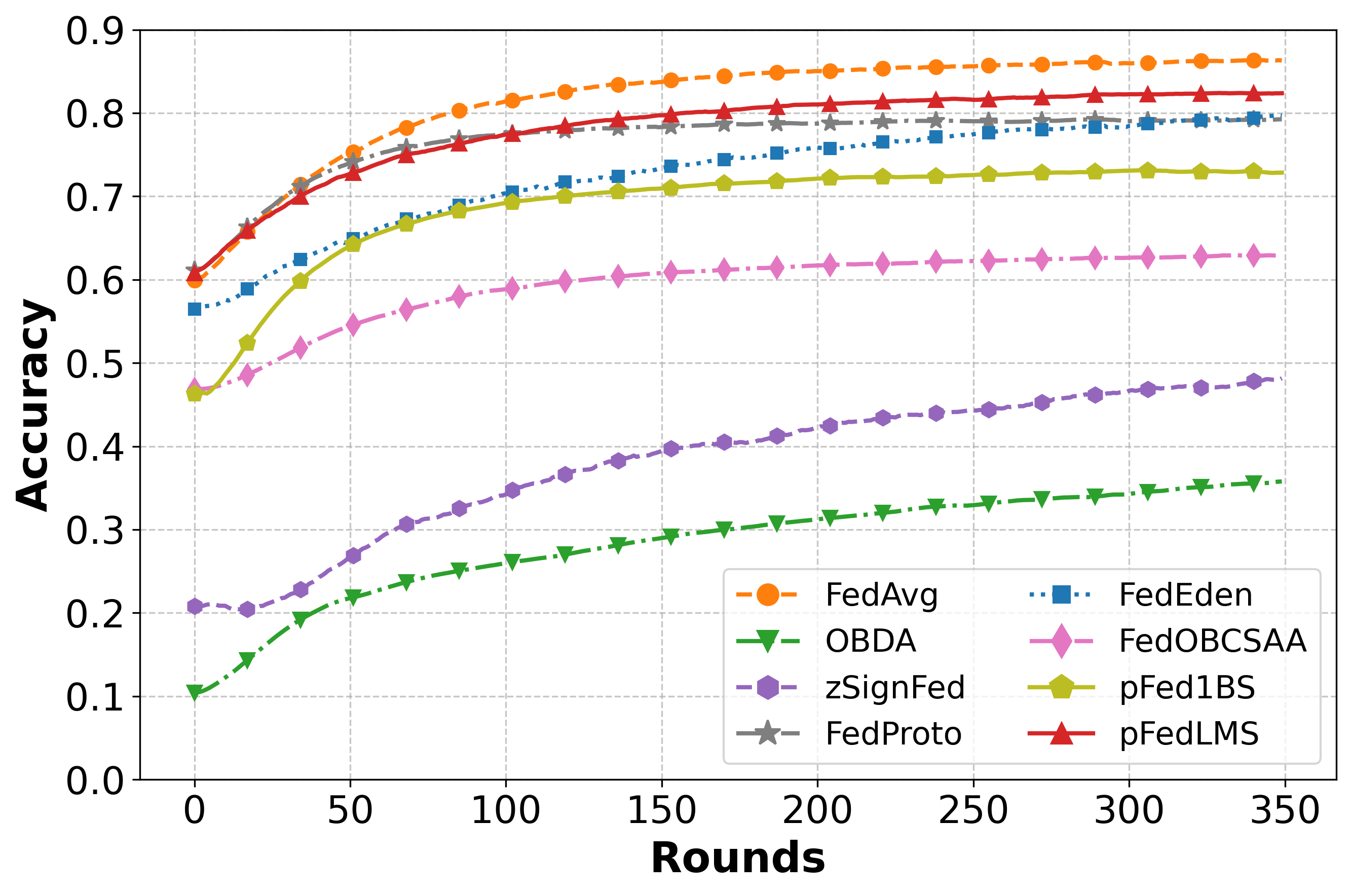}
		\caption{Comparison of test accuracy versus communication rounds for different algorithms under non-IID settings $\alpha=0.5$ on the CIFAR-10 dataset.}
		\label{fig:all_accuracy}
	\end{figure}
  	\begin{figure*}[!t]
		\centering
		\includegraphics[width=0.75\textwidth]{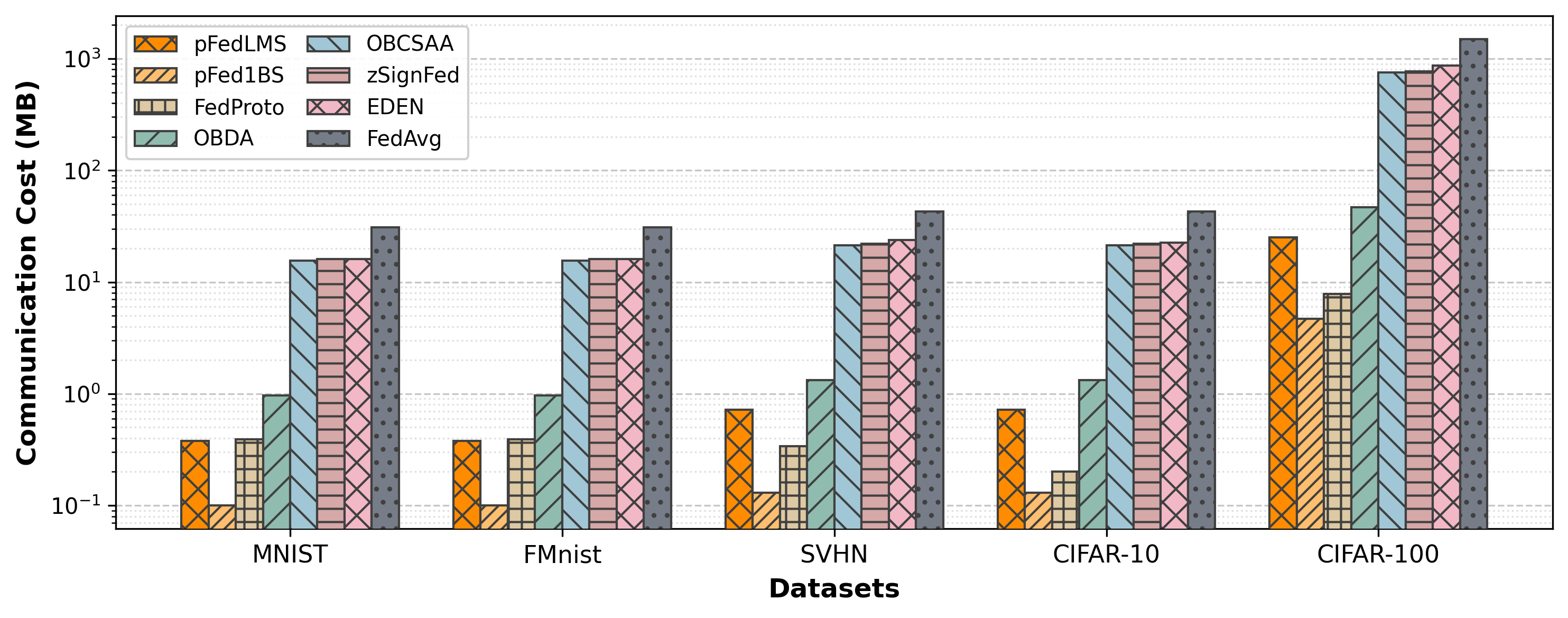} 
		\caption{Comparison of communication cost for different algorithms across five datasets.}
		\label{fig:communication_cost}
	\end{figure*}  
The test accuracy results across five datasets under two distinct Dirichlet non-IID settings are summarized in Tables~\ref{tab:main_accuracy_results_alpha05} and \ref{tab:main_accuracy_results_alpha01}. The proposed pFedLMS demonstrates competitive performance against communication-efficient baselines across most scenarios while closely approaching the full-precision FedAvg baseline on several datasets. Specifically, when $\alpha=0.5$, pFedLMS achieves high accuracy on all datasets and clearly outperforms most communication-efficient baselines. The advantage is especially evident on CIFAR-10 and CIFAR-100, where pFedLMS improves over pFed1BS by 9.17\% and 13.49\%, respectively. This indicates that the layer-wise multi-threshold design provides a more informative consensus signal than the layer-agnostic one-bit sketching strategy. When $\alpha=0.1$, pFedLMS remains robust under stronger data heterogeneity. It achieves the best result on FMNIST among the communication-efficient baselines and obtains competitive performance on MNIST, SVHN, and CIFAR-10. On CIFAR-100, pFedLMS substantially improves over all communication-efficient baselines, outperforming pFed1BS by 10.85\%. These results show that the proposed method can preserve personalized accuracy while using compact bidirectional low-bit sketches.


Fig.~\ref{fig:all_accuracy} shows the convergence curves on CIFAR-10 under the non-IID setting with $\alpha=0.5$. FedAvg obtains the highest final accuracy, but it relies on full-precision model transmission. In contrast, pFedLMS achieves the best performance among the compressed methods and approaches FedAvg with much lower communication cost. Compared with pFed1BS, FedEDEN, OBDA, zSignFed, and FedOBCSAA, pFedLMS maintains a consistently higher curve throughout most of the training process. This result indicates that the layer-wise multi-threshold consensus provides more effective guidance for local model updates than single-threshold or sign-based compressed messages.


\begin{figure}
    \centering
    \includegraphics[width=0.9\linewidth]{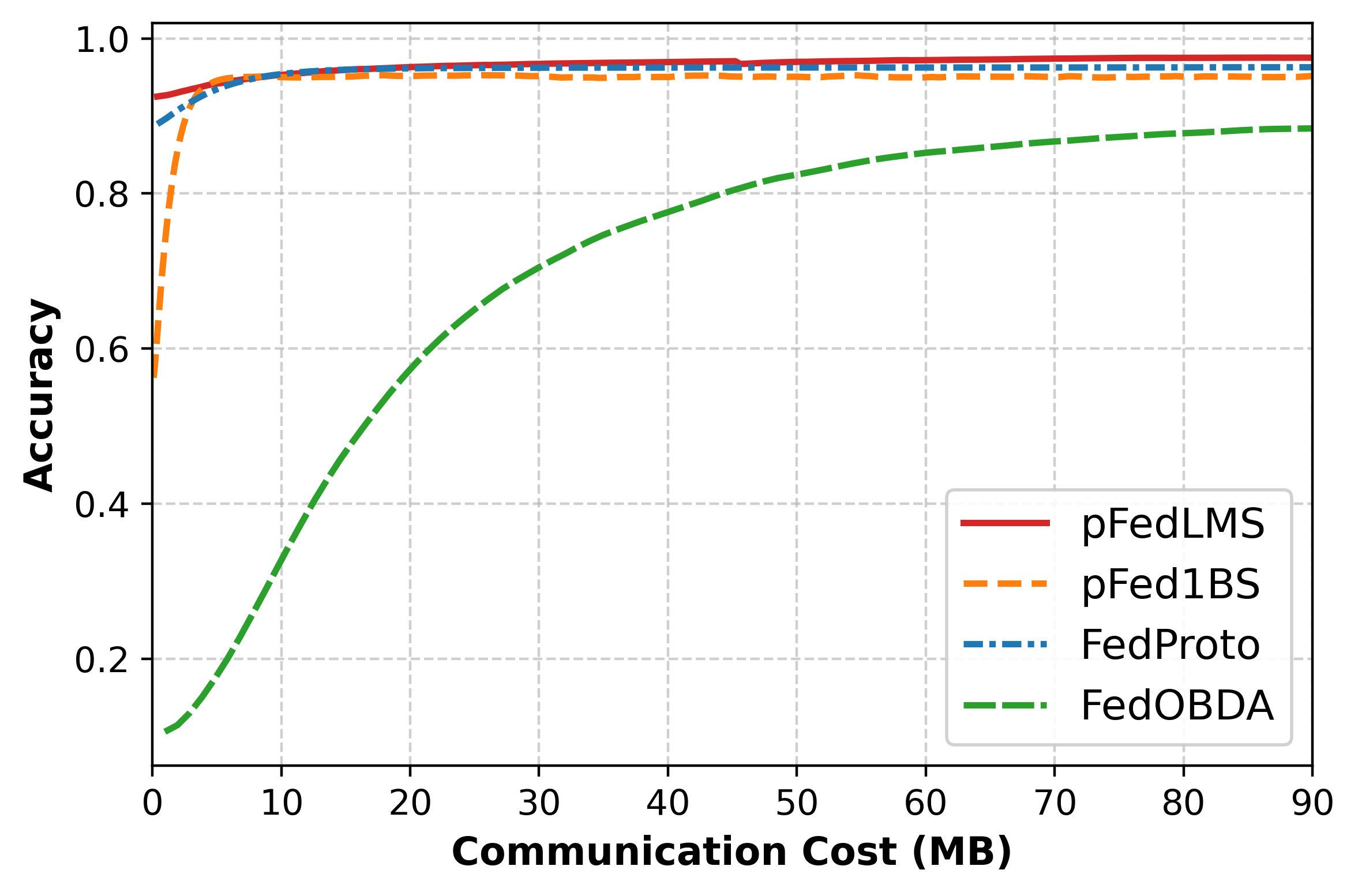}
    \caption{Comparison of test accuracy against cumulative communication cost among low-communication algorithms on the MNIST dataset.}
    \label{fig:communication_accuracy}
\end{figure}

To highlight the transmission efficiency, Fig.~\ref{fig:communication_cost} reports the cumulative communication cost of all methods on a logarithmic scale. FedAvg incurs the largest overhead, requiring $31.06$ MB on MNIST/FMNIST, $42.68$ MB on SVHN/CIFAR-10, and $1495.34$ MB on CIFAR-100. In contrast, pFedLMS requires only $0.38$ MB on MNIST/FMNIST, $0.72$ MB on SVHN/CIFAR-10, and $25.05$ MB on CIFAR-100, corresponding to communication cost reductions of approximately $98.8\%$, $98.3\%$, and $98.3\%$ compared with FedAvg, respectively. Compared with EDEN and zSignFed, pFedLMS also achieves a much lower communication cost while maintaining higher accuracy. Although FedProto and pFed1BS can be highly compact in some cases, their performance drops on complex datasets such as CIFAR-100. These results show that pFedLMS provides a favorable balance between accuracy and communication efficiency.

Fig.~\ref{fig:communication_accuracy} further illustrates the test accuracy as a function of cumulative communication cost on the MNIST dataset. The proposed pFedLMS achieves the best overall curve among the compared low-communication methods. It reaches a high accuracy at the very beginning and quickly stabilizes above $95\%$ with only a small communication budget. In comparison, pFed1BS and FedProto also converge rapidly, but their final accuracies are slightly lower than that of pFedLMS. FedOBDA exhibits a much slower growth pattern: its accuracy is still around $40\%$ at about $10$ MB and increases gradually as the communication cost grows. These results show that pFedLMS can obtain higher accuracy with fewer transmitted bits, confirming the effectiveness of layer-wise multi-threshold sketching under bandwidth-constrained settings.

\subsection{Ablation Study}

To evaluate the effectiveness of the key components in the pFedLMS framework, we conduct ablation experiments on both the layer-wise sketching design and the multi-threshold quantization.

\begin{table}[ht!]
\centering
\renewcommand{\arraystretch}{1.2}
\setlength{\tabcolsep}{6pt}
\textbf{(a) SVHN Dataset} \\
\vspace{1mm}
\resizebox{\linewidth}{!}{
\begin{tabular}{lccc}
\toprule
\textbf{Method} & \textbf{Multi-threshold} & \textbf{Layer-wise} & \textbf{Acc. (\%)}\\
\midrule
Refined pFed1BS & \xmark & \xmark & 89.80 \\ 
pFedLMS ($T=1$) & \xmark & \cmark & 90.31\\
only Multi-threshold & \cmark & \xmark & 91.41\\
pFedLMS ($T=7$) & \cmark & \cmark & 91.94\\
\bottomrule
\end{tabular}
}

\vspace{0.4cm}

\textbf{(b) CIFAR-10 Dataset} \\
\vspace{1mm}
\resizebox{\linewidth}{!}{
\begin{tabular}{lccc}
\toprule
\textbf{Method} & \textbf{Multi-threshold} & \textbf{Layer-wise} & \textbf{Acc. (\%)}\\
\midrule
Refined pFed1BS & \xmark & \xmark & 78.68 \\ 
pFedLMS ($T=1$) & \xmark & \cmark & 79.89 \\
only Multi-threshold & \cmark & \xmark & 81.17\\
pFedLMS ($T=7$) & \cmark & \cmark & 82.61 \\
\bottomrule
\end{tabular}
}
\caption{Ablation results for layer-wise and multi-threshold designs on different datasets.}
\label{tab:ablation_settings}
\end{table}

Table \ref{tab:ablation_settings} compares the performance of four algorithmic variants on the SVHN and CIFAR-10 datasets. Here, the refined pFed1BS baseline is obtained by removing both the layer-wise design and the multi-threshold mechanism from pFedLMS. It differs from the original pFed1BS in that the threshold is the $0.5$-quantile of the sketched values rather than zero. The results show that both the layer-wise design and the multi-threshold mechanism contribute to the final performance. On SVHN, the refined pFed1BS baseline achieves an accuracy of $89.80\%$. Introducing layer-wise sketching with $T=1$ improves the accuracy to $90.31\%$, while using the multi-threshold mechanism alone further raises it to $91.41\%$. When the two designs are combined, pFedLMS with $T=7$ reaches $91.94\%$, giving an overall gain of $2.14\%$ over the baseline. A similar trend can be observed on CIFAR-10, where the accuracy increases from $78.68\%$ to $82.61\%$. These results confirm that the two components are complementary and jointly improve the quality of the low-bit consensus.


\noindent
\begin{figure}
    \centering
    \includegraphics[width=0.9\linewidth]{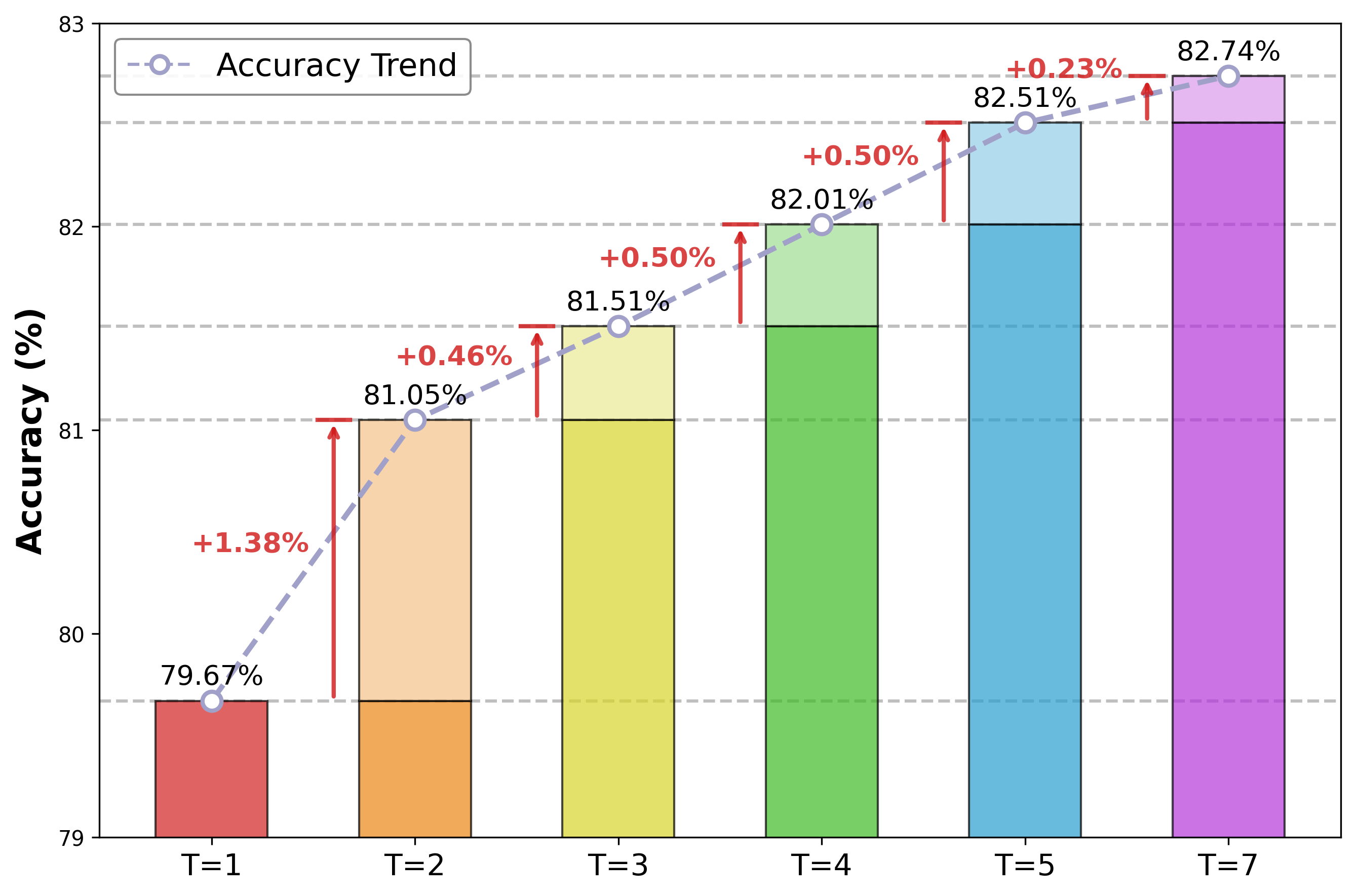}
    \caption{Test accuracy of pFedLMS with different quantization thresholds $T$ on the CIFAR-10 dataset.}
    \label{fig:ablation_t_values}
\end{figure}

To further study the effect of multi-threshold quantization, we evaluate pFedLMS with different numbers of thresholds on the CIFAR-10 dataset, where $T \in \{1,2,3,4,5,7\}$. As shown in Fig.~\ref{fig:ablation_t_values}, the test accuracy increases as $T$ becomes larger. When $T=1$, the method reduces to a sign-based sketching scheme and achieves $79.67\%$ accuracy. Increasing $T$ to $2$ improves the accuracy to $81.05\%$, which gives the largest gain among all tested settings. Further increasing $T$ to $3$, $4$, $5$, and $7$ leads to steady but smaller improvements, and the accuracy finally reaches $82.74\%$. These results indicate that using more thresholds can provide a finer description of the sketched model representation. The gain becomes smaller when $T$ is large, suggesting that a small number of thresholds is already sufficient to capture most of the useful consensus information. Therefore, pFedLMS can improve personalization accuracy while still maintaining low-bit communication.

\section{Conclusion}
In this paper, we proposed pFedLMS, a communication-efficient personalized federated learning framework for edge intelligence with limited bandwidth. The proposed method uses layer-wise multi-threshold random sketching to obtain more expressive low-bit representations than conventional one-bit compression. By combining compact uplink and downlink transmission with a consensus regularizer, pFedLMS reduces communication cost while preserving personalized model training. We also provided a convergence analysis for the resulting nonconvex stochastic optimization problem. Experimental results show that pFedLMS achieves a favorable balance between communication cost and accuracy compared with representative federated learning baselines.

\section*{Acknowledgment}
This work was supported by the National Natural Science Foundation of China under Grant No. 62501432 and U25A20528.

\section*{CRediT authorship contribution statement}

\noindent \textbf{Xu Zhang}: Conceptualization, Writing -- original draft, Methodology, Formal analysis; \textbf{Xingyu Hou}: Methodology, Visualization, Writing -- original draft; \textbf{Jiacheng Cheng}: Methodology, Investigation, Writing -- original draft; \textbf{Kaiyuan Feng}: Visualization, Writing -- Review \& Editing; \textbf{Maoguo Gong}: Supervision, Conceptualization, Writing -- Review \& Editing.




\appendix
	\setcounter{table}{0}
	\section{Communication Protocol} \label{app:comm_details}
	
	In this section, we present the practical communication protocol for the proposed layer-wise multi-threshold random sketching framework. Although the theoretical formulation introduces $T$ one-bit comparison functions per layer, the actual implementation transmits a compact $(T+1)$-ary representation, which is subsequently encoded into a binary bitstream for efficient communication.

	\subsection{Client-Side (T+1)-ary Encoding}
	Recall that for each client $k$ and layer $\ell$, the sketched representation is given by
	\begin{equation}
		\bm{y}_{k,\ell} = \bm{\Phi}_\ell \bm{\theta}_{k,\ell} \in \mathbb{R}^{m_\ell},
	\end{equation}
	with an associated ordered threshold set $\boldsymbol{\tau}_\ell^r = \{\tau_{\ell,1}^r < \cdots < \tau_{\ell,T}^r\}$. The multi-threshold comparisons are defined as
	\begin{equation}
		b_{\ell,t}^k(i) = \mbox{sign} \big( y_{k,\ell}(i) - \tau_{\ell,t}^r \big), \quad t = 1,\dots,T.
	\end{equation}
	
	Instead of transmitting the full collection of $T$ one-bit comparison vectors
	\begin{equation}
		\bm{B}_{k,\ell} = \big( \bm{b}_{\ell,1}^k, \dots, \bm{b}_{\ell,T}^k \big),
	\end{equation}
	the client aggregates these comparisons into a single $(T+1)$-ary symbol at each coordinate. Specifically, for each $i \in \{1,\dots,m_\ell\}$, we define
	\begin{equation}
		q_{k,\ell}(i) = \sum_{t=1}^{T} \mathbb{I}\!\left( b_{\ell,t}^k(i) = 1 \right) \in \{0,1,\dots,T\}.
	\end{equation}
	The value $q_{k,\ell}(i)$ uniquely identifies the interval in which $y_{k,\ell}(i)$ lies.
	
	Collecting all coordinates yields the $(T+1)$-ary sketch vector
	\begin{equation}
		\bm{q}_{k,\ell} = \big( q_{k,\ell}(1), \dots, q_{k,\ell}(m_\ell) \big) \in \{0,1,\dots,T\}^{m_\ell}.
	\end{equation}
	
	To minimize communication overhead, the client jointly encodes $\bm q_{k,\ell}$ as a single nonnegative integer via its base-$(T+1)$ representation:
	\begin{equation}
		Q_{k,\ell} = \sum_{i=1}^{m_\ell} q_{k,\ell}(i)\,(T+1)^{\,i-1}.
	\end{equation}
	The integer $Q_{k,\ell}$ is then converted into a binary bitstream and transmitted to the server. The required number of transmitted bits satisfies
	\begin{equation}
		\left\lceil \log_2\big((T+1)^{m_\ell}\big) \right\rceil = \left\lceil m_\ell \log_2(T+1) \right\rceil,
	\end{equation}
	which achieves the minimum fixed-length binary representation for $(T+1)$-ary symbols.
	
	\begin{remark}
		The base $(T+1)$ encoding above provides a mathematical description of the compact representation and its bit complexity. In practice, the vector $\bm q_{k,\ell}$ is divided into several short blocks rather than encoded as a single large integer. Each block is then packed independently into a binary representation of fixed length. This implementation avoids integer overflow and is compatible with standard communication protocols based on byte alignment. Since the mapping is lossless, it preserves the original $(T+1)$-ary symbols and does not affect the aggregation rule at the server or the regularization at the clients. The total communication cost remains $\lceil m_\ell\log_2(T+1)\rceil$ bits, apart from a small overhead caused by block alignment.
	\end{remark}
	
	\subsection{Server-Side Decoding and Aggregation}
	
	Upon receiving the binary bitstream from client $k$, the server first converts
	the bitstream back into the corresponding nonnegative integer $Q_{k,\ell}$.
	The integer is then decoded into the $(T+1)$-ary sketch vector
	$\bm q_{k,\ell}$ by successive base-$(T+1)$ expansion:
	\begin{equation}
		q_{k,\ell}(i) = \left\lfloor \frac{Q_{k,\ell}}{(T+1)^{\,i-1}} \right\rfloor \bmod (T+1), \quad i = 1,\dots,m_\ell,
	\end{equation}
	where $\bmod$ denotes the modulo operator, i.e., the remainder of integer division. This decoding step exactly recovers the compact $(T+1)$-ary representation without any information loss.
	
	\textbf{Recovery of Comparison Vectors.}
	Although the $(T+1)$-ary symbols are used for communication efficiency, the aggregation rule and the client-side regularization are defined on threshold-wise one-bit comparisons. Therefore, the server deterministically maps each $(T+1)$-ary symbol back to $T$ one-bit comparison results according to
	\begin{equation}
		\bar b_{\ell,t}^k(i) = \mathbb{I}\!\left( q_{k,\ell}(i) \ge t \right), \quad b_{\ell,t}^k(i) = 2\bar b_{\ell,t}^k(i)-1.
	\end{equation}
	This mapping is lossless and recovers the exact comparison outcomes that would have been obtained by directly transmitting the vectors $\{\bm b_{\ell,t}^k\}_{t=1}^T$.
	
	\textbf{Threshold-Wise Aggregation.}
	All clients participate in the current communication round.
	For each layer $\ell$, threshold index $t$, and coordinate $i$, the server computes the aggregated comparison result via weighted majority voting:
	\begin{equation}
		v_{\ell,t}(i) = \mbox{sign} \Big( \sum_{k=1}^K p_k \, b_{\ell,t}^k(i) \Big), \quad t = 1,\dots,T .
	\end{equation}
	The collection $\bm{V} = \{\bm{v}_{\ell,t}\}_{\ell=1,t=1}^{L,T}$ constitutes the layer-wise aggregated comparison vectors, which are identical to those obtained under the original multi-threshold aggregation rule.
	
	\textbf{Re-encoding for Downlink Transmission.}
	To enable efficient downlink communication, the aggregated comparison vectors are reassembled into a \((T+1)\)-ary consensus representation. By Lemma~\ref{lem:interval}, for each layer \(\ell\) and coordinate \(i\), the aggregated sequence \(\{v_{\ell,t}(i)\}_{t=1}^T\) is monotone nonincreasing in \(t\). Therefore, it can be represented losslessly by the number of positive entries:
	\[
	\tilde q_\ell(i)
	=
	\sum_{t=1}^T \mathbb I(v_{\ell,t}(i)=1)
	\in \{0,1,\ldots,T\}.
	\]
	The resulting \((T+1)\)-ary consensus vector \(\tilde{\bm q}_\ell\in\{0,1,\ldots,T\}^{m_\ell}\) is then encoded as
	\[
	\tilde Q_\ell
	=
	\sum_{i=1}^{m_\ell}
	\tilde q_\ell(i)(T+1)^{i-1},
	\]
	converted into a binary bitstream, and broadcast to all clients.
	
	\begin{remark}
		The above decoding and aggregation procedure ensures that the server-side aggregation semantics remain unchanged. The $(T+1)$-ary representation is introduced purely as a compact communication format and does not alter the threshold-wise aggregation rule or the client-side regularization.
	\end{remark}
	
	\subsection{Client-Side Decoding}
	
	Upon receiving the binary bitstream broadcast by the server, each client first reconstructs the corresponding nonnegative integer $\tilde{Q}_{\ell}$ and decodes it into the $(T+1)$-ary consensus vector $\tilde{\bm q}_{\ell}$ via base-$(T+1)$ expansion:
	\begin{equation}
		\tilde{\bm q}_{\ell}(i) = \left\lfloor \frac{\tilde{Q}_{\ell}}{(T+1)^{\,i-1}} \right\rfloor \bmod (T+1), \quad i = 1,\dots,m_\ell .
	\end{equation}
	
	\textbf{Recovery of Threshold-Wise Consensus.}
	Since the client-side regularization is defined on threshold-wise consensus vectors, the decoded \((T+1)\)-ary symbols are mapped back to \(T\) one-bit consensus vectors by
	\[
	\bar v_{\ell,t}(i)
	=
	\mathbb I(\tilde q_\ell(i)\ge t),
	\quad
	v_{\ell,t}(i)
	=
	2\bar v_{\ell,t}(i)-1,
	\quad t=1,\ldots,T.
	\]
	The recovered collection \(\bm V=\{\bm v_{\ell,t}\}_{\ell=1,t=1}^{L,T}\) is then used in the multi-threshold alignment regularizer and its smoothed gradient for local model updates.
	
	\begin{remark}
		The above decoding procedure exactly recovers the same threshold-wise consensus information as in the original formulation. Therefore, the proposed $(T+1)$-ary encoding and binary transmission do not alter
		the client-side regularization or optimization dynamics.
	\end{remark}
	
	\subsection{Communication Complexity}
	
	\textbf{Uplink Communication.}
	For each client $k$ and layer $\ell$, the transmitted message is the binary representation of the base-$(T+1)$ encoded integer $Q_{k,\ell}$. Since $Q_{k,\ell} \in \{0,\dots,(T+1)^{m_\ell}-1\}$, the number of transmitted bits per layer satisfies
	\begin{equation}
		\mathrm{Bits}_{\ell}^{\mathrm{up}} = \left\lceil m_\ell \log_2(T+1) \right\rceil .
	\end{equation}
	Therefore, the total uplink communication cost per round per client is
	\begin{equation}
		\mathrm{Bits}^{\mathrm{up}} = \sum_{\ell=1}^{L} \left\lceil m_\ell \log_2(T+1) \right\rceil .
	\end{equation}
	
	\textbf{Downlink Communication.}
	The server aggregates the received sketches and encodes the layer-wise consensus $\tilde{\bm q}_{\ell}$ using the same base-$(T+1)$ representation. The downlink message thus requires
	\begin{equation}
		\mathrm{Bits}_{\ell}^{\mathrm{down}} = \left\lceil m_\ell \log_2(T+1) \right\rceil
	\end{equation}
	bits per layer. Since the same consensus is broadcast to all clients, the total downlink communication cost per round is
	\begin{equation}
		\mathrm{Bits}^{\mathrm{down}} = \sum_{\ell=1}^{L} \left\lceil m_\ell \log_2(T+1) \right\rceil.
	\end{equation}

	\section{Hyperparameter Sensitivity Analysis}
	
	\subsection{Effect of Sketching Operator Refresh Interval}

	We study the effect of the sketching operator refresh interval on MNIST. The sketching operators are refreshed every $h$ communication rounds, where $h \in \{1,2,3,5,10,20,50\}$. A smaller $h$ corresponds to more frequent refreshes, while a larger $h$ keeps the same operators for more rounds. All other settings are identical to those used in the main experiments.

	As shown in Table~\ref{tab:refresh_frequency}, more frequent refreshes generally lead to better performance. The highest accuracy of $97.85\%$ is achieved when the sketching operators are refreshed every round. As $h$ increases from $1$ to $50$, the accuracy gradually decreases to $96.44\%$. This result suggests that frequently changing the sketching directions provides more diverse measurements and improves the quality of the global consensus. Although less frequent refreshes can reduce objective variation across rounds, the reduced sketch diversity leads to lower empirical accuracy in this experiment.
	
	\begin{table}[!t] \centering   \small \setlength{\tabcolsep}{3pt} \begin{tabular}{c|ccccccc} \hline \shortstack{$h$} & 1 & 2 & 3 & 5 & 10 & 20 & 50 \\ \hline Accuracy (\%) & \textbf{97.85} & 97.73 & 97.57 & 97.34 & 97.00 & 96.61 & 96.44 \\ \hline \end{tabular} 
    \caption{Effect of the sketching operator refresh interval $h$ on MNIST. The best result is shown in bold.}
    \label{tab:refresh_frequency}
	\end{table} 
	\subsection{Sensitivity to the Regularization Parameter}
	
    We further evaluate the sensitivity of pFedLMS to the regularization parameter $\lambda$. Table~\ref{tab:lambda_ablation} reports the results on five datasets under $\alpha=0.1$ and $T=7$, with $\lambda \in \{0.2,0.1,0.05,0.01,$ $0.005\}$ and all other hyperparameters unchanged. Overall, our method pFedLMS remains relatively stable across a broad range of $\lambda$, particularly on MNIST, FMNIST, and CIFAR-10. Specifically, $\lambda=0.005$ achieves the best performance on MNIST and FMNIST, $\lambda=0.01$ performs best on SVHN and CIFAR-100, and $\lambda=0.1$ gives the highest accuracy on CIFAR-10. The results also indicate that an excessively small $\lambda$ may weaken global consensus alignment, as observed on SVHN and CIFAR-10, whereas a relatively large $\lambda$ may introduce stronger consensus drift and reduce performance on MNIST, FMNIST, and CIFAR-100. Therefore, a moderate or relatively small value of $\lambda$ generally provides a favorable balance between collaborative alignment and optimization stability.

	\begin{table}[t]
		\centering
		 \small \setlength{\tabcolsep}{5pt}
		\begin{tabular}{c|ccccc}
			\hline
			$\lambda$ & MNIST & FMNIST & SVHN & CIFAR-10 & CIFAR-100 \\
		\hline
		0.2   & 98.25 & 97.09 & 95.75 & 93.60 & 65.59 \\
		0.1   & 98.54 & 97.36 & 95.29 & \textbf{93.65} & 65.22 \\
		0.05  & 98.65 & 97.42 & 95.78 & 93.56 & 66.25 \\
		0.01  & 98.83 & 97.63 & \textbf{95.85} & 93.43 & \textbf{66.84} \\
		0.005 & \textbf{98.85} & \textbf{97.64} & 94.92 & 93.21 & 66.14 \\
		\hline
		\end{tabular}
        \caption{Effect of the regularization parameter $\lambda$ under $\alpha=0.1$ and $T=7$. The best result in each column is shown in bold.}
        \label{tab:lambda_ablation}
	\end{table}

	\section{Proof of Theorems and Lemmas}
	
	\subsection{Proof of Theorem~\ref{thm:convergence}}
	
	For notational simplicity, define
	\begin{equation}
		S_r \triangleq \sum_{k=1}^K p_k \sum_{e=0}^{E-1} \mathbb E\!\left[ \left\| \nabla F_k(\bm{\theta}_k^{r,e};\bm V^r) \right\|_2^2 \right].
		\label{eq:Sr-definition}
	\end{equation}
	
	By summing the one-step descent bound in Lemma~\ref{lem:one-step} over the local steps $e=0,\ldots,E-1$ and then over all clients with weights $\{p_k\}_{k=1}^K$, we obtain
	\begin{align}
		&\sum_{k=1}^K p_k \mathbb E\!\left[ F_k(\bm{\theta}_k^{r+1};\bm V^r) \right] \nonumber\\
		&\le \sum_{k=1}^K p_k \mathbb E\!\left[ F_k(\bm{\theta}_k^{r};\bm V^r) \right] - \frac{\eta}{2}S_r + \frac{E L_F\eta^2}{2}\sigma^2 \nonumber\\
		&= \mathbb E[\Psi^r] - \frac{\eta}{2}S_r + \frac{E L_F\eta^2}{2}\sigma^2.
		\label{eq:round-descent-proof}
	\end{align}
	
	We next account for the change of the objective between two consecutive communication rounds. By Lemma~\ref{lem:threshold_update_aware_drift}, for every client $k$,
	\begin{equation}
		\mathbb E\!\left[ \left| F_k(\bm{\theta}_k^{r+1};\bm V^{r+1}) - F_k(\bm{\theta}_k^{r+1};\bm V^r) \right| \right] \le \lambda\beta^r.
		\label{eq:client-drift-proof}
	\end{equation}
	Therefore,
	\begin{align}
		\mathbb E\!\left[ F_k(\bm{\theta}_k^{r+1};\bm V^{r+1}) \right] \le \mathbb E\!\left[ F_k(\bm{\theta}_k^{r+1};\bm V^r) \right] + \lambda\beta^r.
		\label{eq:client-drift-one-sided}
	\end{align}
	Multiplying \eqref{eq:client-drift-one-sided} by $p_k$, summing over $k=1,\ldots,K$, and using $\sum_{k=1}^K p_k=1$, we obtain
	\begin{align}
		\mathbb E[\Psi^{r+1}] &= \sum_{k=1}^K p_k \mathbb E\!\left[ F_k(\bm{\theta}_k^{r+1};\bm V^{r+1}) \right] \nonumber\\
		&\le \sum_{k=1}^K p_k \mathbb E\!\left[ F_k(\bm{\theta}_k^{r+1};\bm V^r) \right] + \lambda\beta^r.
		\label{eq:global-drift-proof}
	\end{align}
	
	Combining \eqref{eq:round-descent-proof} and \eqref{eq:global-drift-proof} gives the following recursion:
	\begin{equation}
		\mathbb E[\Psi^{r+1}] \le \mathbb E[\Psi^r] - \frac{\eta}{2}S_r + \frac{E L_F\eta^2}{2}\sigma^2 + \lambda\beta^r.
		\label{eq:potential-recursion-proof}
	\end{equation}
	
	Summing \eqref{eq:potential-recursion-proof} over $r=0,\ldots,R-1$ yields
	\begin{align}
		\mathbb E[\Psi^R] \le \mathbb E[\Psi^0] - \frac{\eta}{2} \sum_{r=0}^{R-1}S_r + \frac{R E L_F\eta^2}{2}\sigma^2 + \lambda \sum_{r=0}^{R-1}\beta^r.
		\label{eq:potential-telescoping-proof}
	\end{align}
	
	By Assumption~\ref{ass:lower}, $\Psi^R\ge\Psi_{\inf}$. Therefore, $\mathbb E[\Psi^R]\ge\Psi_{\inf}$, and \eqref{eq:potential-telescoping-proof} implies
	\begin{align}
		\frac{\eta}{2} \sum_{r=0}^{R-1}S_r \le& \mathbb E[\Psi^0]-\Psi_{\inf} + \frac{R E L_F\eta^2}{2}\sigma^2 + \lambda \sum_{r=0}^{R-1}\beta^r.
		\label{eq:sum-gradient-bound-proof}
	\end{align}
	
	Dividing both sides of \eqref{eq:sum-gradient-bound-proof} by $\eta E R/2$ gives
	\begin{align}
		&\frac{1}{R} \sum_{r=0}^{R-1} \frac{1}{E} \sum_{e=0}^{E-1} \sum_{k=1}^K p_k \mathbb E\!\left[ \left\| \nabla F_k(\bm{\theta}_k^{r,e};\bm V^r) \right\|_2^2 \right] \nonumber\\
		&\le \frac{2\big(\mathbb E[\Psi^0]-\Psi_{\inf}\big)}{\eta E R} + L_F\eta\sigma^2 + \frac{2\lambda}{\eta E R} \sum_{r=0}^{R-1}\beta^r.
		\label{eq:main-convergence-proof}
	\end{align}
	
	This completes the proof.
	
	\subsection{Proof of Lemma \ref{lem:majority_voting}}
	Recall that for any $a\in\{\pm1\}$ and $v\in\{\pm1\}$,
	\begin{equation}
		2[va]_{-}=1-va.
	\end{equation}
	Applying this identity elementwise gives
	\begin{equation}
		\left\| \left[ \bm v_{\ell,t}\odot\bm b_{\ell,t}^{k} \right]_{-} \right\|_1 = \frac{1}{2} \left( m_\ell- \left\langle \bm v_{\ell,t},\bm b_{\ell,t}^{k} \right\rangle \right).
	\end{equation}
	Since $\sum_{k=1}^{K}p_k=1$, the terms involving $m_\ell$ are independent of $\bm V$. Therefore, Problem~\eqref{eq:server_aggregation} is equivalent to
	\begin{equation}
		\max_{\bm V\in\mathcal V} \sum_{\ell=1}^{L}\sum_{t=1}^{T} \left\langle \bm v_{\ell,t}, \sum_{k=1}^{K}p_k\bm b_{\ell,t}^{k} \right\rangle.
	\end{equation}
	The resulting problem is separable across layers, thresholds, and coordinates. For each coordinate $i$, an optimal solution satisfies
	\begin{equation}
		v_{\ell,t}^{\star}(i) \in \arg\max_{v\in\{\pm1\}} v\sum_{k=1}^{K}p_k b_{\ell,t}^{k}(i).
	\end{equation}
	Hence,
	\begin{equation}
		\bm v_{\ell,t}^{\star} = \operatorname{sign} \left( \sum_{k=1}^{K}p_k\bm b_{\ell,t}^{k} \right).
	\end{equation}
	If the weighted sum is zero at a coordinate, both signs are optimal, and the convention $\operatorname{sign}(0)=1$ selects one of them. This completes the proof.
	
	\subsection{Proof of Lemma \ref{lem:interval}}
	
	Since $\tau_{\ell,1}<\cdots<\tau_{\ell,T}$, for each client $k$ the sequence $\{\mbox{\rm sign}(y_k-\tau_{\ell,t})\}_{t=1}^T$ is monotone nonincreasing in $t$. Because $p_k\ge 0$, the weighted sum
	\[
	\sum_{k=1}^K p_k\,\mbox{\rm sign}(y_k-\tau_{\ell,t})
	\]
	is also monotone nonincreasing in $t$. Together with the definition of $v_{\ell,t}(i)$
	\[
	v_{\ell,t}(i)
	=
	\mbox{\rm sign}\!\left(
	\sum_{k=1}^K p_k\,\mbox{\rm sign}(y_k-\tau_{\ell,t})
	\right),
	\]
	we get that the aggregated sequence $\{v_{\ell,t}(i)\}_{t=1}^T$ is monotone nonincreasing, and there exists $s_{\ell,i}\in\{0,1,\dots,T\}$ such that it changes from $+1$ to $-1$ at $s_{\ell,i}$.
	
	For the coordinate-wise penalty in \eqref{eq:reg} to be zero, it is necessary and sufficient that
	\[
	v_{\ell,t}(i)
	\big(
	(\bm{\Phi}_\ell \bm{\theta}_{k,\ell})(i)-\tau_{\ell,t}
	\big)
	\ge 0,
	\quad t=1,\ldots,T.
	\]
	Using the sign pattern above, these inequalities are equivalent to
	\[
	(\bm{\Phi}_\ell \bm{\theta}_{k,\ell})(i)\ge \tau_{\ell,t},
	\quad t\le s_{\ell,i},
	\]
	and
	\[
	(\bm{\Phi}_\ell \bm{\theta}_{k,\ell})(i)\le \tau_{\ell,t},
	\quad t>s_{\ell,i}.
	\]
	Since the thresholds are ordered, this is equivalent to
	\[
	\tau_{\ell,s_{\ell,i}}
	\le
	(\bm{\Phi}_\ell \bm{\theta}_{k,\ell})(i)
	\le
	\tau_{\ell,s_{\ell,i}+1}.
	\]
	This completes the proof.
	
	\subsection{Proof of Lemma \ref{lem:resolution_gain}}
	If $s_{\ell,i}\in\{1,\ldots,T-1\}$, then Lemma~\ref{lem:interval} gives
	\[
	\mathcal I_{\ell,i}
	=
	[
	\tau_{\ell,s_{\ell,i}},
	\tau_{\ell,s_{\ell,i}+1}
	].
	\]
	Therefore,
	\[
	|\mathcal I_{\ell,i}|
	=
	\tau_{\ell,s_{\ell,i}+1}
	-
	\tau_{\ell,s_{\ell,i}}
	\le
	\Delta_{\ell,T}.
	\]
	
	If the thresholds are selected as quantiles $q_t=t/(T+1)$, then each internal interval $[\tau_{\ell,t},\tau_{\ell,t+1}]$ for $t=1,\ldots,T-1$ carries probability mass $1/(T+1)$. Hence,
	\[
	\frac{1}{T+1}
	=
	\int_{\tau_{\ell,t}}^{\tau_{\ell,t+1}}
	f_\ell^{\rm ref}(z)\,dz
	\ge
	f_{\ell,\min}^{\rm ref}
	\left(
	\tau_{\ell,t+1}-\tau_{\ell,t}
	\right).
	\]
	Thus,
	\[
	\tau_{\ell,t+1}-\tau_{\ell,t}
	\le
	\frac{1}{(T+1)f_{\ell,\min}^{\rm ref}}.
	\]
	Taking the maximum over $t=1,\ldots,T-1$ gives the desired result.

	\subsection{Proof of Lemma \ref{lem:bounded-model}}
	
	Based on \eqref{eq:g_hat}, define
	\begin{equation}
		\widehat{\bm g}_k^{r,e}
		\triangleq
		\widehat{\bm g}_k(\bm{\theta}_k^{r,e};\mathcal B_k^{r,e}),
		\quad
		\bm{h}_k^{r,e}
		\triangleq
		\nabla \widetilde{\mathcal R}_{\rho}(\bm{\theta}_k^{r,e};\bm{V}^r).
	\end{equation}
	Then the client update can be written as
	\begin{equation}
		\bm{\theta}_k^{r,e+1}
		=
		(1-\eta\mu)\bm{\theta}_k^{r,e}
		-\eta\big(\widehat{\bm g}_k^{r,e}+\lambda \bm{h}_k^{r,e}\big).
		\label{eq:bounded-proof-update}
	\end{equation}

	We first bound the regularizer gradient uniformly.
	From the block expression of $\nabla \widetilde{\mathcal R}_{\rho}$, for each layer $\ell$,
	\begin{multline}
		\nabla_{\bm{\theta}_\ell}\widetilde{\mathcal R}_{\rho}(\bm{\theta};\bm{V}^r) \\
		= \frac{1}{T} (\bm{\Phi}_\ell^r)^\top \sum_{t=1}^T \left[ \mathrm{clip}_{[-1,1]} \!\left( \frac{\bm{\Phi}_\ell^r\bm{\theta}_\ell-\tau_{\ell,t}^r\mathbf 1}{\rho} \right) - \bm{v}_{\ell,t}^r \right].
	\end{multline}
	For each $t$, every coordinate of the vector
	\[
	\mathrm{clip}_{[-1,1]}\!\left(
	\frac{\bm{\Phi}_\ell^r\bm{\theta}_\ell-\tau_{\ell,t}^r\mathbf 1}{\rho}
	\right)-\bm v_{\ell,t}^r
	\]
	lies in $[-2,2]$, because the clipping term belongs to $[-1,1]$ and $\bm v_{\ell,t}^r\in\{\pm1\}^{m_\ell}$. Hence,
	\begin{equation}
		\left\| \mathrm{clip}_{[-1,1]} \!\left( \frac{\bm{\Phi}_\ell^r\bm{\theta}_\ell-\tau_{\ell,t}^r\mathbf 1}{\rho} \right) - \bm{v}_{\ell,t}^r \right\|_2 \le 2\sqrt{m_\ell}.
	\end{equation}
	Using the triangle inequality and Assumption~\ref{ass:sketch}, we obtain
	\begin{align}
		&\left\| \nabla_{\bm{\theta}_\ell}\widetilde{\mathcal R}_{\rho}(\bm{\theta};\bm{V}^r) \right\|_2 \nonumber\\
		&\le \frac{1}{T} \|(\bm{\Phi}_\ell^r)^\top\|_2 \sum_{t=1}^T \left\| \mathrm{clip}_{[-1,1]} \!\left( \frac{\bm{\Phi}_\ell^r\bm{\theta}_\ell-\tau_{\ell,t}^r\mathbf 1}{\rho} \right) - \bm{v}_{\ell,t}^r \right\|_2 \nonumber\\
		&\le \frac{1}{T}\kappa_\ell \sum_{t=1}^T 2\sqrt{m_\ell} \nonumber\\
		&= 2\kappa_\ell \sqrt{m_\ell}. 
		\label{eq:block-reg-grad-bound}
	\end{align}
	Stacking all layer blocks gives
	\begin{align}
		\|\nabla \widetilde{\mathcal R}_{\rho}(\bm{\theta};\bm{V}^r)\|_2^2 \!=& \sum_{\ell=1}^L \left\| \nabla_{\bm{\theta}_\ell}\widetilde{\mathcal R}_{\rho}(\bm{\theta};\bm{V}^r) \right\|_2^2 \nonumber\\
		\le& \sum_{\ell=1}^L 4\kappa_\ell^2 m_\ell \triangleq C_R^2. 
		\label{eq:reg-grad-global-bound}
	\end{align}
	Therefore,
	\begin{equation}
		\|\bm{h}_k^{r,e}\|_2 \le C_R, \quad \forall k,r,e. 
		\label{eq:h-bound}
	\end{equation}
	
	Next, taking squared norms on both sides of \eqref{eq:bounded-proof-update}, we have
	\begin{align}
		&\|\bm{\theta}_k^{r,e+1}\|_2^2 = \left\| (1-\eta\mu)\bm{\theta}_k^{r,e} - \eta(\widehat{\bm g}_k^{r,e}+\lambda\bm{h}_k^{r,e}) \right\|_2^2 \nonumber\\
		&= (1-\eta\mu)^2\|\bm{\theta}_k^{r,e}\|_2^2 -2\eta(1-\eta\mu) \left\langle \bm{\theta}_k^{r,e}, \widehat{\bm g}_k^{r,e}+\lambda\bm{h}_k^{r,e} \right\rangle \nonumber\\
		&\quad+\eta^2\|\widehat{\bm g}_k^{r,e}+\lambda\bm{h}_k^{r,e}\|_2^2. 
		\label{eq:expand-theta}
	\end{align}
	Using Young's inequality,
	\begin{equation}
		\begin{aligned}
			2\left| \left\langle \bm{\theta}_k^{r,e}, \widehat{\bm g}_k^{r,e}+\lambda\bm{h}_k^{r,e} \right\rangle \right| \le \mu\|\bm{\theta}_k^{r,e}\|_2^2 + \frac{1}{\mu} \|\widehat{\bm g}_k^{r,e}+\lambda\bm{h}_k^{r,e}\|_2^2,
		\end{aligned}
	\end{equation}
	and substituting it into \eqref{eq:expand-theta}, we obtain
	\begin{align}
		\|\bm{\theta}_k^{r,e+1}\|_2^2 &\le \big((1-\eta\mu)^2+\eta\mu(1-\eta\mu)\big) \|\bm{\theta}_k^{r,e}\|_2^2 \nonumber\\
		&\quad+ \left( \frac{\eta(1-\eta\mu)}{\mu}+\eta^2 \right) \|\widehat{\bm g}_k^{r,e}+\lambda\bm{h}_k^{r,e}\|_2^2 \nonumber\\
		&= (1-\eta\mu)\|\bm{\theta}_k^{r,e}\|_2^2 \nonumber\\
		&\quad+ \left( \frac{\eta(1-\eta\mu)}{\mu}+\eta^2 \right) \|\widehat{\bm g}_k^{r,e}+\lambda\bm{h}_k^{r,e}\|_2^2 \nonumber\\
		&= (1-\eta\mu)\|\bm{\theta}_k^{r,e}\|_2^2 +  \frac{\eta}{\mu} \|\widehat{\bm g}_k^{r,e}+\lambda\bm{h}_k^{r,e}\|_2^2. 
		\label{eq:theta-recursion-pre}
	\end{align}
	Applying $\|\bm{a}+\bm{b}\|_2^2\le 2\|\bm{a}\|_2^2+2\|\bm{b}\|_2^2$ gives
	\begin{equation}
		\begin{aligned}
			\|\widehat{\bm g}_k^{r,e}+\lambda\bm{h}_k^{r,e}\|_2^2 \le 2\|\widehat{\bm g}_k^{r,e}\|_2^2 + 2\lambda^2\|\bm{h}_k^{r,e}\|_2^2.
		\end{aligned} 
		\label{eq:sum-square-bound}
	\end{equation}
	Taking expectation on both sides of \eqref{eq:theta-recursion-pre}, and then using
	Assumption~\ref{ass:task-grad}, \eqref{eq:h-bound}, and \eqref{eq:sum-square-bound}, we get
	\begin{align}
		\mathbb{E}\!\left[\|\bm{\theta}_k^{r,e+1}\|_2^2\right] \le& (1-\eta\mu)\mathbb{E}\!\left[\|\bm{\theta}_k^{r,e}\|_2^2\right] \nonumber\\
		&+ \frac{2\eta}{\mu} \left( \mathbb{E}\!\left[\|\widehat{\bm g}_k^{r,e}\|_2^2\right] + \lambda^2\mathbb{E}\!\left[\|\bm{h}_k^{r,e}\|_2^2\right] \right) \nonumber\\
		&\le (1-\eta\mu)\mathbb{E}\!\left[\|\bm{\theta}_k^{r,e}\|_2^2\right] \nonumber\\
		&\quad+ \frac{2\eta}{\mu} \left( G^2+\lambda^2 C_R^2 \right) \nonumber\\
		&= (1-\eta\mu)\mathbb{E}\!\left[\|\bm{\theta}_k^{r,e}\|_2^2\right] + C_B. 
		\label{eq:theta-recursion-final}
	\end{align}
	
	Now define
	$x_{r,e}^{(k)} \triangleq
	\mathbb{E}\!\left[\|\bm{\theta}_k^{r,e}\|_2^2\right], \alpha \triangleq 1-\eta\mu.$
	Since $0< \eta \le 1/\mu$, we have $0 \le \alpha < 1$.
	Thus \eqref{eq:theta-recursion-final} becomes
	\begin{equation}
		x_{r,e+1}^{(k)} \le \alpha x_{r,e}^{(k)} + C_B. 
		\label{eq:scalar-recursion}
	\end{equation}
	By induction over all local steps and all communication rounds,
	\begin{equation}
		x_{r,e}^{(k)} \le \max\left\{ \mathbb{E}\|\bm{\theta}_k^0\|_2^2,\, \frac{C_B}{\eta\mu} \right\}. 
		\label{eq:second-moment-bound}
	\end{equation}
	Taking the maximum over $k=1,\dots,K$, we obtain
	\begin{equation}
		\begin{aligned}
			\mathbb{E}\!\left[\|\bm{\theta}_k^{r,e}\|_2^2\right] \le& \max\left\{ \max_{1\le k\le K} \mathbb{E}\|\bm{\theta}_k^0\|_2^2,\, \frac{C_B}{\eta\mu} \right\} = B_{\theta}^2, \forall k,r,e.
		\end{aligned} 
		\label{eq:uniform-second-moment}
	\end{equation}
	This completes the proof.
	
	\subsection{Proof of Lemma \ref{lem:F-smooth}}
	
	We prove the result by bounding the Lipschitz continuity of the gradient of each term in $F_k$.
	By Assumption \ref{ass:smooth}, $f_k$ is $L_{f}$-smooth, hence
	\begin{equation}
		\begin{aligned}
			\|\nabla f_k( \bm \theta_k)-\nabla f_k(\bm \theta_k')\| \le& L_{f}\|\bm \theta_k- \bm \theta_k'\|.
		\end{aligned}
	\end{equation}
	
	Recall that the gradient of $\widetilde{\mathcal R}_{\rho}$ with respect to the $\ell$-th layer parameter $\bm \theta_{k,\ell}$ is given by
	\begin{multline}
		\nabla_{\theta_{k,\ell}} \widetilde{\mathcal R}_{\rho}(\bm \theta_k; \bm V^r) \\
		= \frac{1}{T}\left[\bm \Phi_\ell^r \right]^{\top} \sum_{t=1}^{T} \left( \mathrm{clip}_{[-1,1]} \!\left( \frac{\bm \Phi_\ell^r \bm \theta_{k,\ell}-\tau_{\ell,t}^r\mathbf{1}}{\rho} \right) - \bm v_{\ell,t}^r \right).
	\end{multline}
	Let $\bm \theta_{k,\ell}, \bm \theta'_{k,\ell}$ be arbitrary. Using the fact that the element-wise clipping operator $\mathrm{clip}_{[-1,1]}(\cdot)$ is $1$-Lipschitz, we obtain
	\begin{equation}
		\begin{aligned}
			&\big\| \nabla_{\bm \theta_{k,\ell}} \widetilde{\mathcal R}_{\rho}(\bm \theta_k; \bm V^r) - \nabla_{\bm \theta_{k,\ell}} \widetilde{\mathcal R}_{\rho}(\bm \theta_k'; \bm V^r) \big\| \\
			\le& \frac{1}{T}\|\left[\bm \Phi_\ell^r \right]^{\top}\| \sum_{t=1}^{T} \bigg{\|} \mathrm{clip}_{[-1,1]}\!\left( \frac{\bm \Phi_\ell^r \bm \theta_{k,\ell}-\tau_{\ell,t}^r\mathbf{1}}{\rho} \right) \\
			&- \mathrm{clip}_{[-1,1]}\!\left( \frac{\bm \Phi_\ell^r \bm \theta'_{k,\ell}-\tau_{\ell,t}^r\mathbf{1}}{\rho} \right) \bigg\| \\
			\le& \frac{1}{\rho}\|\bm \Phi_\ell^r\|^2 \,\|\bm \theta_{k,\ell}-\bm \theta'_{k,\ell}\| \\
			=& \frac{\kappa_\ell^2}{\rho} \|\bm \theta_{k,\ell}-\bm \theta'_{k,\ell}\|.
		\end{aligned}
	\end{equation}
	Summing over layers yields
	\begin{multline}
		\big\| \nabla_{\bm \theta_{k}} \widetilde{\mathcal R}_{\rho}(\bm \theta_k; \bm V^r) - \nabla_{\bm \theta_{k}} \widetilde{\mathcal R}_{\rho}(\bm \theta_k'; \bm V^r) \big\| \\
		\le \max_{\ell} \frac{\kappa_\ell^2}{\rho} \,\|\bm \theta_{k}-\bm \theta'_{k}\|.
	\end{multline}
	Note that the gradient of $\frac{\mu}{2}\|\bm \theta_k\|_2^2$ is $\mu \bm \theta_k$, which is $\mu$-Lipschitz. Combining the above bounds and using the linearity of the gradient, we conclude that $F_k(\cdot;\bm{V}^r)$ is $L_F$-smooth with
	\begin{equation}
		\begin{aligned}
			L_F =& L_{f} + \max_{\ell}\frac{\lambda \kappa_\ell^2}{\rho} + \mu.
		\end{aligned}
	\end{equation}

	\subsection{Proof of Lemma \ref{lem:one-step}}
	
	For brevity, write
	\begin{equation}
		\widehat{\bm g}_{k}^{r,e}
		\triangleq
		\widehat{\bm g}_{k}(\bm{\theta}_k^{r,e};\mathcal B_k^{r,e}).
	\end{equation}
	Define the stochastic update direction as
	\begin{equation}
		\bm G_k^{r,e}
		\triangleq
		\widehat{\bm g}_{k}^{r,e}
		+\lambda\nabla_{\bm{\theta}_k}\widetilde{\mathcal R}_{\rho}
		(\bm{\theta}_k^{r,e};\bm V^r)
		+\mu\bm{\theta}_k^{r,e}.
		\label{eq:G_def}
	\end{equation}
	Then the client update is
	\begin{equation}
		\bm{\theta}_{k}^{r,e+1}-\bm{\theta}_{k}^{r,e}
		=
		-\eta \bm G_k^{r,e}.
	\end{equation}
	Moreover, by the definition of $F_k$, we have
	\begin{equation}
		\nabla F_k(\bm{\theta}_{k}^{r,e};\bm V^r)
		=
		\nabla f_k(\bm{\theta}_{k}^{r,e})
		+\lambda\nabla_{\bm{\theta}_k}\widetilde{\mathcal R}_{\rho}
		(\bm{\theta}_k^{r,e};\bm V^r)
		+\mu\bm{\theta}_k^{r,e}.
		\label{eq:F_grad_def}
	\end{equation}
	
	By Lemma~\ref{lem:F-smooth} and the standard descent lemma,
	\begin{align}
		F_k(\bm{\theta}_{k}^{r,e+1};\bm V^r)
		\le\,&
		F_k(\bm{\theta}_{k}^{r,e};\bm V^r) \nonumber\\
		&+
		\left\langle
		\nabla F_k(\bm{\theta}_{k}^{r,e};\bm V^r),
		\bm{\theta}_{k}^{r,e+1}-\bm{\theta}_{k}^{r,e}
		\right\rangle \nonumber\\
		&+
		\frac{L_F}{2}
		\|\bm{\theta}_{k}^{r,e+1}-\bm{\theta}_{k}^{r,e}\|_2^2 .
	\end{align}
	Substituting $\bm{\theta}_{k}^{r,e+1}-\bm{\theta}_{k}^{r,e}
	=-\eta\bm G_k^{r,e}$ gives
	\begin{multline}
		F_k(\bm{\theta}_{k}^{r,e+1};\bm V^r)
		\le
		F_k(\bm{\theta}_{k}^{r,e};\bm V^r) \\
		-\eta
		\left\langle
		\nabla F_k(\bm{\theta}_{k}^{r,e};\bm V^r),
		\bm G_k^{r,e}
		\right\rangle
		+
		\frac{L_F\eta^2}{2}
		\|\bm G_k^{r,e}\|_2^2 .
		\label{eq:descent_with_G}
	\end{multline}
	
	Taking the conditional expectation with respect to the mini-batch randomness and using Assumption~\ref{ass:sgd}, we obtain
	\begin{equation}
		\mathbb{E}\!\left[
		\bm G_k^{r,e}
		\,\middle|\,
		\bm{\theta}_{k}^{r,e}
		\right]
		=
		\nabla F_k(\bm{\theta}_{k}^{r,e};\bm V^r).
		\label{eq:G_unbiased}
	\end{equation}
	Therefore,
	\begin{multline}
		\mathbb{E}\!\left[
		F_k(\bm{\theta}_{k}^{r,e+1};\bm V^r)
		\,\middle|\,
		\bm{\theta}_{k}^{r,e}
		\right]
		\le
		F_k(\bm{\theta}_{k}^{r,e};\bm V^r) \\
		-\eta
		\|\nabla F_k(\bm{\theta}_{k}^{r,e};\bm V^r)\|_2^2
		+
		\frac{L_F\eta^2}{2}
		\mathbb{E}\!\left[
		\|\bm G_k^{r,e}\|_2^2
		\,\middle|\,
		\bm{\theta}_{k}^{r,e}
		\right].
		\label{eq:conditional_descent}
	\end{multline}
	
	From \eqref{eq:G_def} and \eqref{eq:F_grad_def}, we have
	\begin{equation}
		\bm G_k^{r,e}
		=
		\nabla F_k(\bm{\theta}_{k}^{r,e};\bm V^r)
		+
		\left(
		\widehat{\bm g}_{k}^{r,e}
		-
		\nabla f_k(\bm{\theta}_{k}^{r,e})
		\right).
	\end{equation}
	By Assumption~\ref{ass:sgd},
	\begin{equation}
		\mathbb{E}\!\left[
		\|\bm G_k^{r,e}\|_2^2
		\,\middle|\,
		\bm{\theta}_{k}^{r,e}
		\right]
		\le
		\|\nabla F_k(\bm{\theta}_{k}^{r,e};\bm V^r)\|_2^2
		+
		\sigma^2 .
		\label{eq:G_second_moment}
	\end{equation}
	Substituting \eqref{eq:G_second_moment} into \eqref{eq:conditional_descent} gives
	\begin{multline}
		\mathbb{E}\!\left[
		F_k(\bm{\theta}_{k}^{r,e+1};\bm V^r)
		\,\middle|\,
		\bm{\theta}_{k}^{r,e}
		\right]
		\le
		F_k(\bm{\theta}_{k}^{r,e};\bm V^r) \\
		-\eta\left(1-\frac{L_F\eta}{2}\right)
		\|\nabla F_k(\bm{\theta}_{k}^{r,e};\bm V^r)\|_2^2
		+
		\frac{L_F\eta^2}{2}\sigma^2 .
	\end{multline}
	When $\eta\le 1/L_F$, we have
	$1-L_F\eta/2\ge 1/2$, which completes the proof.

	\subsection{Proof of Lemma~\ref{lem:threshold_update_aware_drift}}
	
	Since $f_k(\bm{\theta}_k^{r+1})$ and $\frac{\mu}{2}\|\bm{\theta}_k^{r+1}\|_2^2$ are identical in $F_k(\bm{\theta}_k^{r+1};\bm V^{r+1})$ and $F_k(\bm{\theta}_k^{r+1};\bm V^r)$, it suffices to bound the variation of the regularization term:
	\begin{multline}
		F_k(\bm{\theta}_k^{r+1};\bm V^{r+1})
		-
		F_k(\bm{\theta}_k^{r+1};\bm V^r) \\
		=
		\lambda\left(
		\widetilde{\mathcal R}_{\rho}
		(\bm{\theta}_k^{r+1};\bm V^{r+1})
		-
		\widetilde{\mathcal R}_{\rho}
		(\bm{\theta}_k^{r+1};\bm V^r)
		\right).
	\end{multline}
	
	For brevity, define
	\begin{equation}
		\bm a_{\ell,t}^r
		\triangleq
		\bm{\Phi}_\ell^r\bm{\theta}_{k,\ell}^{r+1}
		-
		\tau_{\ell,t}^r\mathbf 1,
		\quad
		\bm a_{\ell,t}^{r+1}
		\triangleq
		\bm{\Phi}_\ell^{r+1}\bm{\theta}_{k,\ell}^{r+1}
		-
		\tau_{\ell,t}^{r+1}\mathbf 1.
	\end{equation}
	By the definition of $\widetilde{\mathcal R}_{\rho}$,
	\begin{multline}
		\widetilde{\mathcal R}_{\rho}
		(\bm{\theta}_k^{r+1};\bm V^{r+1})
		-
		\widetilde{\mathcal R}_{\rho}
		(\bm{\theta}_k^{r+1};\bm V^r) \\
		=
		\frac{1}{T}
		\sum_{\ell=1}^L
		\sum_{t=1}^T
		\Big(
		\|\bm a_{\ell,t}^{r+1}\|_{1,\rho}
		-
		\|\bm a_{\ell,t}^{r}\|_{1,\rho}
		\\-
		\langle
		\bm v_{\ell,t}^{r+1},
		\bm a_{\ell,t}^{r+1}
		\rangle
		+
		\langle
		\bm v_{\ell,t}^{r},
		\bm a_{\ell,t}^{r}
		\rangle
		\Big).
	\end{multline}
	Moreover,
	\begin{align}
		&-\langle
		\bm v_{\ell,t}^{r+1},
		\bm a_{\ell,t}^{r+1}
		\rangle
		+
		\langle
		\bm v_{\ell,t}^{r},
		\bm a_{\ell,t}^{r}
		\rangle
		\nonumber\\
		&=
		-\langle
		\bm v_{\ell,t}^{r+1}-\bm v_{\ell,t}^{r},
		\bm a_{\ell,t}^{r+1}
		\rangle
		-
		\langle
		\bm v_{\ell,t}^{r},
		\bm a_{\ell,t}^{r+1}-\bm a_{\ell,t}^{r}
		\rangle .
	\end{align}
	Hence,
	\begin{multline}
		\left|
		F_k(\bm{\theta}_k^{r+1};\bm V^{r+1})
		-
		F_k(\bm{\theta}_k^{r+1};\bm V^r)
		\right| \\
		\le
		\frac{\lambda}{T}
		\sum_{\ell=1}^L
		\sum_{t=1}^T
		\left(
		A_{\ell,t}^r+
		B_{\ell,t}^r+
		C_{\ell,t}^r
		\right),
		\label{eq:coarse-drift-decomposition}
	\end{multline}
	where
	\begin{equation}
		\begin{aligned}
			A_{\ell,t}^r
			&\triangleq
			\left|
			\|\bm a_{\ell,t}^{r+1}\|_{1,\rho}
			-
			\|\bm a_{\ell,t}^{r}\|_{1,\rho}
			\right|,\\
			B_{\ell,t}^r
			&\triangleq
			\left|
			\left\langle
			\bm v_{\ell,t}^{r+1}-\bm v_{\ell,t}^{r},
			\bm a_{\ell,t}^{r+1}
			\right\rangle
			\right|,\\
			C_{\ell,t}^r
			&\triangleq
			\left|
			\left\langle
			\bm v_{\ell,t}^{r},
			\bm a_{\ell,t}^{r+1}-\bm a_{\ell,t}^{r}
			\right\rangle
			\right|.
		\end{aligned}
	\end{equation}
	
	We first bound $A_{\ell,t}^r$ and $C_{\ell,t}^r$. Since the smoothed $\ell_1$ norm is $1$-Lipschitz with respect to the $\ell_1$ norm,
	\begin{equation}
		A_{\ell,t}^r
		\le
		\|\bm a_{\ell,t}^{r+1}-\bm a_{\ell,t}^{r}\|_1.
	\end{equation}
	Moreover, since
	$\bm v_{\ell,t}^{r}\in\{\pm1\}^{m_\ell}$,
	\begin{equation}
		C_{\ell,t}^r
		\le
		\|\bm a_{\ell,t}^{r+1}-\bm a_{\ell,t}^{r}\|_1.
	\end{equation}
	Therefore,
	\begin{equation}
		A_{\ell,t}^r+C_{\ell,t}^r
		\le
		2\|\bm a_{\ell,t}^{r+1}-\bm a_{\ell,t}^{r}\|_1.
		\label{eq:AC-coarse}
	\end{equation}
	
	By the definitions of $\bm a_{\ell,t}^{r}$ and $\bm a_{\ell,t}^{r+1}$,
	\begin{equation}
		\bm a_{\ell,t}^{r+1}-\bm a_{\ell,t}^{r}
		=
		(\bm{\Phi}_\ell^{r+1}-\bm{\Phi}_\ell^r)
		\bm{\theta}_{k,\ell}^{r+1}
		-
		(\tau_{\ell,t}^{r+1}-\tau_{\ell,t}^{r})\mathbf 1.
	\end{equation}
	Thus,
	\begin{align}
		&\|\bm a_{\ell,t}^{r+1}-\bm a_{\ell,t}^{r}\|_1
		\nonumber\\
		&\le
		\sqrt{m_\ell}
		\|\bm{\Phi}_\ell^{r+1}-\bm{\Phi}_\ell^r\|_2
		\|\bm{\theta}_{k,\ell}^{r+1}\|_2
		+
		m_\ell
		|\tau_{\ell,t}^{r+1}-\tau_{\ell,t}^{r}|.
		\label{eq:affine-difference-bound}
	\end{align}
	
	By Assumption~\ref{ass:sketch} and the definition of
	$\chi_\ell^r$,
	\begin{equation}
		\|\bm{\Phi}_\ell^{r+1}-\bm{\Phi}_\ell^r\|_2
		\le
		2\kappa_\ell\chi_\ell^r.
	\end{equation}
	Moreover, Lemma~\ref{lem:bounded-model} and Jensen's inequality give
	\begin{equation}
		\mathbb E
		\|\bm{\theta}_{k,\ell}^{r+1}\|_2
		\le
		\left(
		\mathbb E
		\|\bm{\theta}_{k,\ell}^{r+1}\|_2^2
		\right)^{1/2}
		\le
		B_\theta.
	\end{equation}
	Assumption~\ref{ass:threshold} further implies
	\begin{equation}
		|\tau_{\ell,t}^{r+1}-\tau_{\ell,t}^{r}|
		\le
		|\tau_{\ell,t}^{r+1}|
		+
		|\tau_{\ell,t}^{r}|
		\le
		2\tau_{\max}.
	\end{equation}
	Taking the expectation in \eqref{eq:AC-coarse} and using \eqref{eq:affine-difference-bound}, we obtain
	\begin{equation}
		\mathbb E\!\left[
		A_{\ell,t}^r+C_{\ell,t}^r
		\right]
		\le
		4\sqrt{m_\ell}\kappa_\ell B_\theta\chi_\ell^r
		+
		4m_\ell\tau_{\max}.
		\label{eq:AC-coarse-final}
	\end{equation}
	
	We next directly bound $B_{\ell,t}^r$. Since $\bm v_{\ell,t}^{r+1},\bm v_{\ell,t}^{r} \in\{\pm1\}^{m_\ell}$,
	\begin{equation}
		\|\bm v_{\ell,t}^{r+1}-\bm v_{\ell,t}^{r}\|_\infty
		\le 2.
	\end{equation}
	Therefore,
	\begin{equation}
		B_{\ell,t}^r
		\le
		2\|\bm a_{\ell,t}^{r+1}\|_1.
	\end{equation}
	Using the definition of $\bm a_{\ell,t}^{r+1}$, we have
	\begin{align}
		\mathbb E[B_{\ell,t}^r]
		&\le
		2\sqrt{m_\ell}
		\mathbb E\!\left[
		\|\bm{\Phi}_\ell^{r+1}
		\bm{\theta}_{k,\ell}^{r+1}\|_2
		\right]
		+
		2m_\ell
		\mathbb E
		|\tau_{\ell,t}^{r+1}|
		\nonumber\\
		&\le
		2\sqrt{m_\ell}\kappa_\ell B_\theta
		+
		2m_\ell\tau_{\max}.
		\label{eq:B-coarse-final}
	\end{align}
	
	Combining \eqref{eq:AC-coarse-final} and
	\eqref{eq:B-coarse-final} yields
	\begin{align}
		&\mathbb E\!\left[
		A_{\ell,t}^r+
		B_{\ell,t}^r+
		C_{\ell,t}^r
		\right] \nonumber\\
		&\le
		4\sqrt{m_\ell}\kappa_\ell B_\theta\chi_\ell^r
		+
		2\sqrt{m_\ell}\kappa_\ell B_\theta
		+
		6m_\ell\tau_{\max} \nonumber\\
		&= 
		\beta_\ell^r.
	\end{align}
	Since the above bound is independent of $t$, averaging over $t=1,\ldots,T$ and summing over $\ell=1,\ldots,L$ in
	\eqref{eq:coarse-drift-decomposition} gives
	\begin{align}
		\mathbb E\!\left[
		\left|
		F_k(\bm{\theta}_k^{r+1};\bm V^{r+1})
		-
		F_k(\bm{\theta}_k^{r+1};\bm V^r)
		\right|
		\right]
		\le
		\lambda
		\sum_{\ell=1}^L
		\beta_\ell^r
		=
		\lambda\beta^r.
	\end{align}
	This completes the proof.









\end{document}